\documentclass[lettersize,journal]{IEEEtran}
\usepackage{amsmath,amsfonts,amssymb}
\usepackage{algorithmic}
\usepackage{algorithm}
\usepackage{array}
\usepackage{multirow}
\usepackage{booktabs}
\usepackage[caption=false,font=normalsize,labelfont=sf,textfont=sf]{subfig}
\usepackage{textcomp}
\usepackage{stfloats}
\usepackage{url}
\usepackage{verbatim}
\usepackage{tcolorbox}
\tcbuselibrary{listings,skins,breakable}
\definecolor{rsclaworange}{RGB}{195,74,20}
\definecolor{rsclawgreen}{RGB}{81,179,51}
\definecolor{rsclawblue}{RGB}{101,200,237}
\definecolor{promptbg}{RGB}{252,250,248}
\lstdefinestyle{promptstyle}{
  basicstyle=\ttfamily\footnotesize,
  breaklines=true,
  columns=fullflexible,
  keepspaces=true,
  showstringspaces=false,
}
\lstdefinestyle{kitstyle}{
  basicstyle=\ttfamily\scriptsize,
  breaklines=false,
  columns=fixed,
  keepspaces=true,
  showstringspaces=false,
}
\newtcblisting{kitbox}[1][]{
  listing only, listing style=kitstyle,
  colback=promptbg, colbacktitle=rsclawgreen,
  coltitle=white, fonttitle=\bfseries\small,
  title=#1, boxrule=0.4pt, colframe=rsclawgreen,
  arc=2pt, left=4pt, right=4pt, top=4pt, bottom=4pt,
  toptitle=3pt, bottomtitle=3pt, titlerule=0pt, breakable,
}
\newtcblisting{promptbox}[1][]{
  listing only, listing style=promptstyle,
  colback=promptbg, colbacktitle=rsclaworange,
  coltitle=white, fonttitle=\bfseries\small,
  title=#1, boxrule=0.4pt, colframe=rsclaworange,
  arc=2pt, left=4pt, right=4pt, top=4pt, bottom=4pt,
  toptitle=3pt, bottomtitle=3pt, titlerule=0pt, breakable,
}
\newtcblisting{rsclawbox}[1][]{
  listing only, listing style=promptstyle,
  colback=promptbg, colbacktitle=rsclawgreen,
  coltitle=white, fonttitle=\bfseries\small,
  title=#1, boxrule=0.4pt, colframe=rsclawgreen,
  arc=2pt, left=4pt, right=4pt, top=4pt, bottom=4pt,
  toptitle=3pt, bottomtitle=3pt, titlerule=0pt, breakable,
}
\newtcolorbox{rsclawframe}[1][]{
  colback=promptbg, colbacktitle=rsclawgreen,
  coltitle=white, fonttitle=\bfseries\small,
  title=#1, boxrule=0.4pt, colframe=rsclawgreen,
  arc=2pt, left=4pt, right=4pt, top=4pt, bottom=4pt,
  toptitle=3pt, bottomtitle=3pt, titlerule=0pt, breakable,
}
\usepackage{graphicx}
\usepackage{cite}
\usepackage[colorlinks=true,linkcolor=blue,citecolor=blue,urlcolor=blue]{hyperref}
\begin{document}

\title{RS-Claw: Progressive Active Tool Exploration via Hierarchical Skill Trees for Remote Sensing Agents}

\author{Liangtian Liu, Zeyuan Wang, Ziyu Li, Kai Ouyang, Zichao Tang,
Chengfu Liu, Haifeng Li,
Hanwen Yu, Wentao Yang, Cheng Yang,
and Dongyang Hou%
\thanks{\textit{(Corresponding authors: Cheng Yang and Dongyang Hou.)}}
\thanks{Liangtian Liu, Zeyuan Wang, Ziyu Li, Kai Ouyang, Zichao Tang, Chengfu Liu, Haifeng Li, Cheng Yang, and Dongyang Hou are with the School of Geosciences and Info-Physics, Central South University, Changsha 410083, China (e-mails: llt62t@csu.edu.cn; 255011039@csu.edu.cn; liziyucsu@csu.edu.cn; oykkksk@csu.edu.cn; 2121050207@csu.edu.cn; 8211220119@csu.edu.cn; lihaifeng@csu.edu.cn; ychades@csu.edu.cn; houdongyang1986@csu.edu.cn).}
\thanks{Hanwen Yu is with the School of Resources and Environment, University of Electronic Science and Technology of China, Xian 710071, China (e-mail: yuhanwenxd@gmail.com).}
\thanks{Wentao Yang is with the School of Earth Sciences and Spatial Information Engineering, Hunan University of Science and Technology, Xiangtan 411201, China, and also with the Sanya Institute of Hunan University of Science and Technology, Sanya 572024, China (e-mail: yangwentao8868@126.com).}}

\newcommand{\IEEEPreprintNotice}{This work has been submitted to the IEEE for possible publication. Copyright may be transferred without notice, after which this version may no longer be accessible.}
\makeatletter
\def\ps@IEEEtitlepagestyle{%
\def\@oddhead{\hbox{}\hfil{\@IEEEheaderstyle\itshape\IEEEPreprintNotice}\hfil\hbox{}}\relax
\def\@evenhead{\hbox{}\hfil{\@IEEEheaderstyle\itshape\IEEEPreprintNotice}\hfil\hbox{}}\relax
\let\@oddfoot\@empty
\let\@evenfoot\@empty}
\makeatother

\maketitle

\begin{abstract}
The rise of multi-modal large language model (MLLM) led remote sensing (RS) intelligence to a new paradigm shift, i.e., from ``see'' to ``action'', especially, OpenClaw-style frameworks promise charming ability that autonomously operate massive RS image processing tools to execute complex tasks. Existing RS agents adopt a passive selection paradigm for tool invocation, relying on either full tool registration (Flat) or retrieval-augmented generation (RAG). However, when confronted with the massive and multi-source heterogeneous RS tool ecosystem, such passive mechanisms struggle to dynamically balance ``context load'' and ``toolset completeness'' throughout the task reasoning process, thus exhibiting inherent limitations: full tool registration triggers context space deficits during long-horizon tasks, whereas RAG retrieval leads to the omission of critical tools in essential steps. To overcome these bottlenecks, this paper redefines the tool selection paradigm---arguing that the agent should act as an active explorer within the tool space. Based on this perspective, we propose RS-Claw, a novel RS agent architecture. By leveraging Skill encapsulation technology at the tool end, this architecture hierarchically structures tool descriptions, enabling the agent to execute on-demand sequential decision-making: initially selecting relevant skill branches by reading only tool summaries, then dynamically loading and reading detailed descriptions, and ultimately achieving precise invocation. This active paradigm not only significantly liberates the agent's context space but also effectively ensures the accurate hit rate of critical tools during long-horizon reasoning. Systematic experiments on the Earth-Bench benchmark demonstrate that RS-Claw's active exploration mechanism effectively filters semantic noise and substantially frees up reasoning space (achieving an input token compression ratio of up to 86\%), comprehensively outperforming existing Flat and RAG baselines across various complex reasoning evaluations.
\end{abstract}


\section{Introduction}

\IEEEPARstart{I}{n} recent years, driven by the continuous advancements of large language models (LLMs) in intent comprehension, autonomous decision-making, and tool utilization, researchers have developed LLM-based agents equipped with specialized toolsets to autonomously execute complex reasoning tasks. These agents are capable of parsing natural language instructions and executing tasks through autonomous planning and the invocation of external tools~\cite{wang2023plan,schick2023toolformer,shen2023hugginggpt,wang2024survey}. In the general domain, for instance, agents such as OpenClaw~\cite{openclaw2026} have demonstrated significant potential in accomplishing tasks via dynamic tool invocation. Capitalizing on the autonomous task-execution capabilities of these agents, researchers in the remote sensing (RS) community have leveraged them to process acquired imagery according to diverse customized requirements. For example, RS-Agent~\cite{xu2024rs} achieves the automated execution of RS tasks across various scenarios---including scene classification, visual question answering (VQA), and object counting---through task decomposition and the collaborative scheduling of multiple tools. Furthermore, Earth-Agent~\cite{feng2025earth} integrates domain-specific tools for index inversion, target recognition, and statistical analysis, thereby facilitating multimodal quantitative spatiotemporal reasoning and scientific analysis.

Most existing RS agents adopt the ``full tool registration (Flat)'' paradigm prevalent in the general domain, wherein the functional descriptions of all candidate tools are directly concatenated into the system prompt. However, the complex workflows of RS data processing and the rapid iteration of multi-source heterogeneous data~\cite{chi2016big} necessitate that these agents be equipped with massive and dynamically expanding tool libraries containing hundreds of functions, such as QGIS, GDAL, Google Earth Engine, and Orfeo ToolBox~\cite{gorelick2017google,grizonnet2017orfeo}. Confronted with such ultra-large-scale toolsets, the traditional Flat paradigm inevitably induces a context space bottleneck within LLMs~\cite{qin2023toolllm}, precipitating two critical challenges during the agent's task reasoning:
\begin{enumerate}
\item Constrained reasoning space in long-horizon tasks. Taking the Earth-Bench evaluation benchmark as an example, the full registration of its 104 domain-specific RS tools and their parameter specifications alone consumes upwards of 20k tokens. Although several state-of-the-art LLMs support extended contexts, lengthy tool descriptions inevitably occupy a substantial portion of the context space. This significantly compresses the essential context space required by the agent for storing intermediate state data, conducting multi-step trial-and-error, and executing ReAct reasoning during long-horizon tasks~\cite{langchain2025benchmarking,bai2024longbench,yao2022react}.

\item Semantic confusion within multi-source tool spaces. Given the disparate physical mechanisms and algorithms associated with different sensors, voluminous and semantically similar tool descriptions generate extensive ``semantic noise.'' Consequently, the model suffers from ``attention defocus'' and the ``lost in the middle'' phenomenon, rendering it highly susceptible to ``tool hallucination'' when parsing user intents~\cite{liu2024lost,patil2024gorilla}.
\end{enumerate}

\begin{figure*}[!t]
\centering
\includegraphics[width=\textwidth]{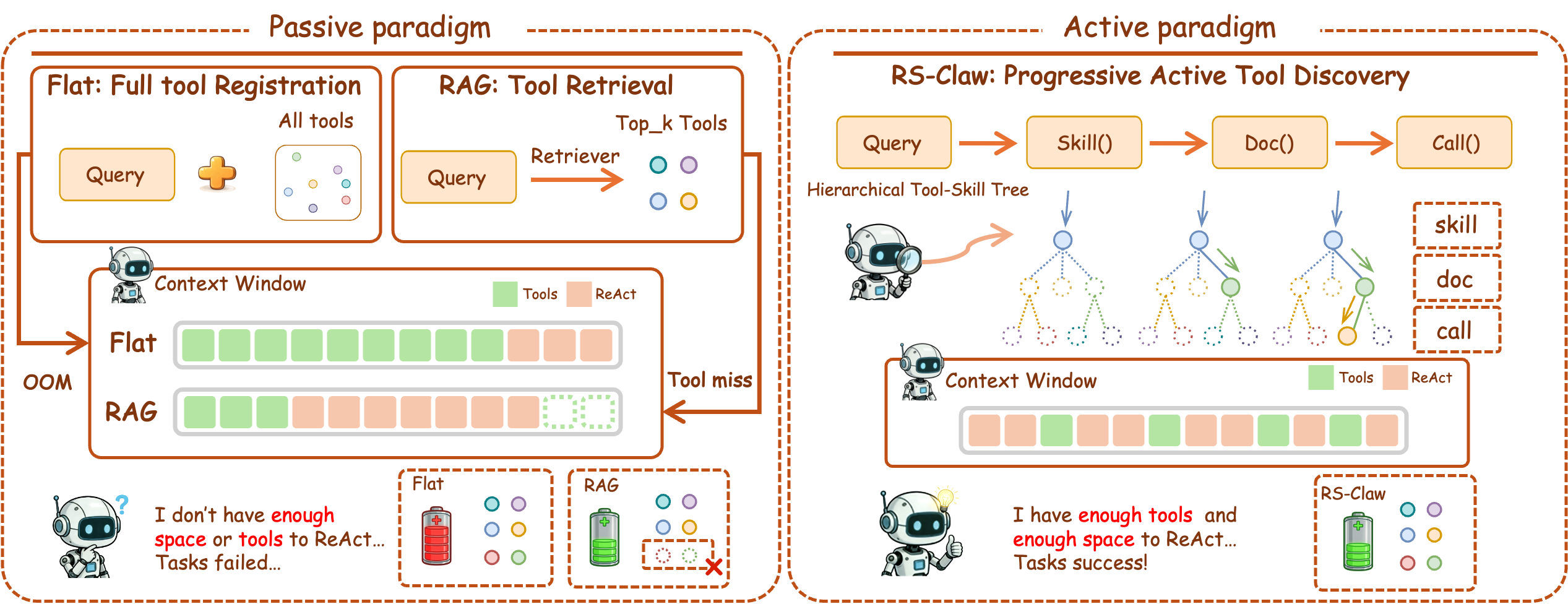}
\caption{Comparison of agent tool selection paradigms. (a)~Passive paradigm: Existing methods define the agent as a passive tool recipient. Specifically, the ``Flat'' strategy globally injects all tools, which leads to context space overflow and severely compresses the reasoning space; the ``RAG'' strategy, despite freeing up some space, is susceptible to causing the omission of critical tools during long-horizon reasoning. (b)~Active paradigm: RS-Claw defines the agent as an active tool explorer. Leveraging a hierarchical skill tree, the agent progressively discovers and loads tools on demand through sequential decision-making. This mechanism, which interleaves dynamic tool loading with ReAct reasoning, simultaneously ensures ample context reasoning space and an accurate tool hit rate.}
\label{fig:paradigm}
\end{figure*}

To mitigate the context bottleneck induced by Flat paradigm, existing research has primarily explored tool registration paradigms based on external retrieval augmentation (RAG)~\cite{lewis2020retrieval}. By filtering a subset of candidate tools from a large-scale tool library prior to task execution, such methods reduce the context space occupied by tool descriptions to a certain extent (e.g., ToolReAgt~\cite{braunschweiler2025toolreagt}). However, this approach still exhibits significant limitations in RS scenarios. RS tasks frequently involve multi-stage task decomposition and state-dependent tool requirements. Relying solely on the initial task description and surface-level semantic similarity for a single-shot retrieval makes it difficult to adequately cover the critical tools required in subsequent reasoning stages. Consequently, although the RAG paradigm alleviates the context load issue, it may sacrifice the completeness of the toolset, making it challenging for the agent to simultaneously balance context efficiency and multi-step reasoning completeness within a large-scale tool space.

Consequently, existing methods still struggle to simultaneously balance exploration efficiency, reasoning completeness, and system scalability within large-scale tool spaces. To break through these bottlenecks, we argue that an agent's tool invocation should mirror human cognitive logic: tool acquisition should neither be a globally static passive retrieval nor a blind code generation process. Instead, it should be a task-driven, active exploration and on-demand discovery process grounded in contextual reasoning. To actualize this mechanism, this paper fundamentally reconstructs the tool-side architecture of the agent, proposing a novel framework named RS-Claw. Inspired by the concept of ``progressive disclosure,'' we discard the traditional flattened tool list. Instead, leveraging expert knowledge, we semantically encapsulate the voluminous RS toolset into ``skills,'' thereby constructing a structured, hierarchical skill tree~\cite{anthropic_agent_skills}. Under this architecture, we innovatively model the discovery and selection of tools as an autonomous sequential decision-making process executed by the agent. Rather than relying on externally predefined program logic for local registration, the agent incorporates ``tool exploration'' as an intrinsic action within its decision space. It is thus guided to conduct progressive, autonomous searches and dynamic loading within the hierarchical tree. This ``active exploration'' paradigm effectively resolves the two aforementioned challenges: First, the on-demand loading of tools reduces the context consumption from a global $O(N)$ to a local $O(K)$, thereby freeing up ample space for the agent to maintain complex intermediate geospatial states and execute multi-step logical reasoning. Second, intent-driven hierarchical exploration filters out extraneous information early in the decision-making pathway, thereby eliminating ``semantic noise'' within the multi-source tool space. This effectively mitigates the issues of attention defocus and tool hallucination.

The main contributions of this article are summarized as follows:
\begin{enumerate}
\item We introduce a pioneering perspective on tool selection for RS agents. Inspired by human cognitive logic, we redefine the tool selection process as an ``active exploration,'' thereby transforming the agent from a passive tool recipient into an active explorer within a structured tool space. Through this theoretical shift, we systematically analyze and address the critical challenges of constrained reasoning space and semantic confusion that agents traditionally encounter under the Flat paradigm.
\item Inspired by the concept of progressive disclosure, we design a hierarchical skill tree at the tool level, innovatively internalizing the tool selection action into the agent's autonomous sequential decision-making process. This architecture enables the agent to dynamically load tool subsets on demand, which not only liberates the reasoning context space but also effectively filters out the semantic confusion and information noise inherent in multi-source toolsets.
\item We conduct systematic experiments on the Earth-Bench benchmark. The proposed RS-Claw comprehensively outperforms both the Flat and RAG baselines across all models and evaluation modes, achieving an improvement of 12.45\% over the Flat baseline in the Qwen3-32b AP mode. In terms of context space optimization, RS-Claw significantly reduces the input token count per round compared to the Flat paradigm, achieving an input token compression ratio of up to 86\% in the Qwen3-32b AP mode. Extensive ablation and scalability experiments further validate the structural rationale and robustness of our proposed method.
\end{enumerate}

\section{Related Work}

Autonomous agents driven by LLMs are progressively reshaping the automation paradigms for scientific computing and Earth observation tasks. Existing research focuses on system architectures with language models as the core controller, integrating multi-step planning, state memory, and tool invocation mechanisms. These systems have demonstrated robust reasoning and decision-making capabilities in complex open-ended tasks, driving the evolution of agents from single-model inference to multi-tool collaborative execution. This paradigm provides a vital foundation for building agents in vertical domains; accordingly, RS tasks have transitioned from multimodal language model adaptation to autonomous systems capable of tool scheduling. However, as the scale of available tools continues to expand, the effective organization and efficient retrieval of large-scale tools during task execution have become critical factors influencing reasoning efficiency and task completion quality. In the context of growing toolsets and increasingly complex task chains, the tool acquisition process still primarily relies on predefined scopes or static single-step decisions, offering limited support for the dynamic evolution of tool requirements during execution. This challenge is particularly pronounced in the field of RS, which is characterized by highly specialized tool ecosystems and complex task dependencies. As RS agents evolve toward increasingly complex long-horizon tasks, the rapid expansion of available tools has further made efficient tool organization and acquisition a critical challenge. To address this issue, our work further focuses on dynamic tool discovery in large-scale tool spaces to support on-demand tool exploration for complex RS tasks.

\subsection{From General-Purpose Agent Systems to RS Task Automation}

In recent years, the paradigm of building autonomous agents with Large Language Models (LLMs) as the core controller has advanced rapidly, becoming a mainstream technical route for complex task automation. Within this framework, the LLM functions as a unified decision-making hub, advancing task execution through sub-goal decomposition, task planning, and the coordination of external tool calls~\cite{wang2024survey}. The ReAct~\cite{yao2022react} framework introduced the interleaved generation of reasoning traces and action decisions, enabling the model to continuously update plans and retrieve external information in dynamic environments, thereby significantly enhancing problem-solving capabilities for complex tasks. Building on this, Reflexion~\cite{shinn2023reflexion} introduced a linguistic self-reflection mechanism, allowing agents to formulate improvement strategies by summarizing past failure trajectories, while Tree of Thoughts~\cite{yao2023tree} studied explicit search over intermediate reasoning paths for complex problem solving.

Driven by this general paradigm, research on task automation in the field of RS has progressively evolved from single-model inference toward Vision-Language Model (VLM) systems geared for multi-task interaction. Early research primarily focused on adapting VLMs to RS scenarios to achieve foundational capabilities such as image captioning, visual question answering (VQA), and spatial understanding. For instance, RSGPT~\cite{hu2025rsgpt} constructed high-quality RS image-text datasets and fine-tuned existing VLMs to enable image description and VQA capabilities. GeoChat~\cite{kuckreja2024geochat} further introduced region-level inputs and spatial coordinate representation mechanisms, allowing the model to perform region-level reasoning and visual grounding, thus supporting more granular spatial interactions. EarthGPT~\cite{zhang2024earthgpt} integrated multi-source sensor data through a unified instruction tuning strategy, achieving unified modeling of multi-task RS interpretation. Building on this, EarthDial~\cite{soni2025earthdial} extended processing to multi-temporal, multi-spectral, and multi-resolution RS data, supporting complex temporal analysis and change detection tasks. RemoteCLIP~\cite{liu2024remoteclip} established a vision-language foundation model pre-trained specifically on RS imagery, providing strong cross-modal alignment for downstream RS tasks, while SkySense~\cite{guo2024skysense} further explored multi-modal RS foundation modeling. However, these systems remain largely instruction-driven and are primarily oriented toward single-step or weakly-planned tasks, showing certain limitations in practical applications that require complex process control and specialized tool invocation.

To enhance RS task automation, researchers have begun exploring RS autonomous agent systems with LLMs as the core controller. Early work such as RS-Agent~\cite{xu2024rs} initially validated the feasibility of LLM-driven RS agents; by using an LLM as the central controller combined with RAG and specialized knowledge bases, it achieved automated execution of RS tasks---including scene classification, visual question answering, and object counting---through task decomposition and multi-tool collaborative scheduling. Subsequently, Earth-Agent~\cite{feng2025earth} integrated 104 domain-specific tools to enable multimodal quantitative spatiotemporal reasoning and scientific analysis. For interactive change analysis, ChangeAgent~\cite{liu2024change} introduced an RS agent framework that combines temporal-image interpretation with user interaction and change reasoning. Addressing task decomposition and cross-domain coordination in complex workflows, GeoLLM-Squad~\cite{lee2025multi} introduced a multi-agent collaboration strategy, where an orchestrator breaks down user requests into sub-tasks for specialized domain agents to process, thereby improving modularity and scalability. Regarding evaluation, Earth-Agent~\cite{feng2025earth} introduced the Earth-Bench benchmark, while GeoLLM-QA~\cite{singh2024evaluating} and ThinkGeo~\cite{shabbir2025thinkgeo} established benchmark settings for tool-augmented agents and real-world RS imagery. Together, these efforts have propelled a paradigm shift in RS agents from single workflows to multi-source scientific analysis.

Although the aforementioned studies have advanced RS tasks from single-step perception toward the automated execution of complex workflows, most existing RS agents still assume that the tool space is predefined and statically accessible---that is, the agent can access the complete set of tools before task execution begins. As the number of specialized tools in the RS domain continues to grow and task pipelines become increasingly long and complex, this fixed tool exposure paradigm gradually faces challenges such as high context load and declining tool selection efficiency. To address this issue, this paper further investigates tool organization and acquisition in large-scale RS tool environments, and explores a more dynamic tool access mechanism that better aligns with the demands of complex tasks.

\subsection{Agent Tool Organization and Retrieval}

As the number of invokable tools for agents continues to grow, the efficient organization and dynamic acquisition of these tools have become critical factors influencing system performance and scalability. Early research typically adopted a Flat paradigm, where all tool descriptions were loaded into the context at once before task execution, allowing the language model to select and invoke them directly. While effective for small-scale toolsets, this approach faces rapidly escalating context load and attention dispersion issues as the number of tools expands to tens of thousands~\cite{qu2025tool,liu2023agentbench}. Consequently, these challenges limit the stability and reasoning efficiency of systems in complex task scenarios. The scale of this challenge has grown substantially: benchmarks such as API-Bank~\cite{li2023api}, ToolBench~\cite{qin2023toolllm}, and Tool Decathlon~\cite{li2025tool} evaluate whether and how LLMs select and invoke external tools across increasingly large-scale tool libraries and long-horizon tasks, while system works such as TaskMatrix.AI~\cite{liang2024taskmatrix} demonstrate modular tool/API orchestration over heterogeneous components.

To alleviate the context load imposed by large-scale tool spaces, researchers have proposed tool selection strategies based on external retrieval mechanisms. These methods employ a retriever to measure the semantic relevance between tool descriptions and sub-tasks, filtering a set of candidate tools from the library for the language model's final decision. For instance, Gorilla~\cite{patil2024gorilla} integrated Retrieval-Aware Training (RAT) into the fine-tuning process to mitigate hallucinations in tool calling. ToolLLM~\cite{qin2023toolllm} constructed ToolBench, a massive benchmark containing over 16,000 real-world REST APIs, and introduced a neural API retriever to support the recall of relevant tools within extensive tool spaces, enabling automated selection for complex tasks. AnyTool~\cite{du2024anytool} implemented a hierarchical API retrieval structure to improve efficiency by narrowing the search scope layer by layer. Re-Invoke~\cite{chen2024re} further studied tool invocation rewriting for zero-shot tool retrieval. Meanwhile, ToolGen~\cite{wang2024toolgen} framed tool selection as a generative decision-making problem, introducing specific tool tokens into the model's vocabulary to achieve unified retrieval and invocation. While these approaches partially relieve context load and significantly boost selection efficiency, they still fundamentally rely on one-shot retrieval. Consequently, they remain prone to overlooking critical tools required for subsequent steps in complex, long-horizon tasks.

Building on these foundations, to address multi-step tool scheduling and dependency issues in complex tasks, researchers have begun exploring more structured and dynamic paradigms for tool organization and retrieval. One representative line of work attempts to organize the tool space by explicitly modeling the execution logic and input-output dependencies between tools. For instance, Voyager~\cite{wang2023voyager}, demonstrated that LLM agents can autonomously build growing skill libraries through lifelong learning in open-ended environments, providing an early precedent for hierarchical skill organization. ToolNet~\cite{liu2024toolnet} moves away from the traditional flat tool list, instead constructing a structured tool network based on state transitions and invocation paths. Through explicit routing mechanisms, it guides the agent to explore the tool space sequentially, thereby enhancing selection efficiency and robustness in large-scale environments. Another approach deeply couples tool retrieval with the reasoning process. For example, ToolReAGt~\cite{braunschweiler2025toolreagt} proposes a reasoning-integrated retrieval-augmented mechanism that combines the agent's intermediate reasoning results with the current context state to dynamically retrieve the most relevant toolsets for subsequent operations, enabling continuous adjustment of the tool selection process. Furthermore, recent studies have begun abstracting tools into reusable skills, building systematic capability organization mechanisms around large-scale skill libraries. Graph of Skills~\cite{li2026graph} proposes a dependency-aware structural retrieval method for large-scale agent skill libraries. Its core idea is to construct executable skill graphs during the offline phase and, during inference, combine semantic-lexical seed retrieval, Personalized PageRank, and context budget constraints to dynamically recall skill sets that include critical prerequisites. This mitigates the issue where pure semantic retrieval might overlook necessary prerequisite skills in complex tasks. In contrast, SkillNet~\cite{liang2026skillnet} focuses on the unified management of agent skills from an infrastructure perspective. By building an open skill network that includes mechanisms for skill creation, evaluation, organization, and connection, it transforms scattered execution experiences into reusable, composable, and evaluable skill assets, supporting long-term capability accumulation across tasks. These studies indicate that agent capability organization is evolving from tool-centric retrieval toward structured management mechanisms oriented around skill dependencies, capability reuse, and long-term evolution.

Although these approaches improve scalability in large-scale tool spaces from perspectives such as semantic retrieval, structured organization, and dynamic code generation, they still fundamentally model tool acquisition as a problem of one-shot retrieval or on-demand generation, lacking an explicit mechanism to capture the evolving nature of tool requirements during task execution. In long-horizon RS tasks, agents often need to continuously adjust subsequent tool selections based on intermediate states, making tool acquisition inherently closer to a process of active exploration paradigm. Motivated by this observation, this paper further formulates tool selection as an active exploration process within a structured tool space, enabling agents to incrementally discover and acquire tools during reasoning rather than making a static decision in a single step.

\section{Method}

\begin{figure*}[!t]
\centering
\includegraphics[width=\textwidth]{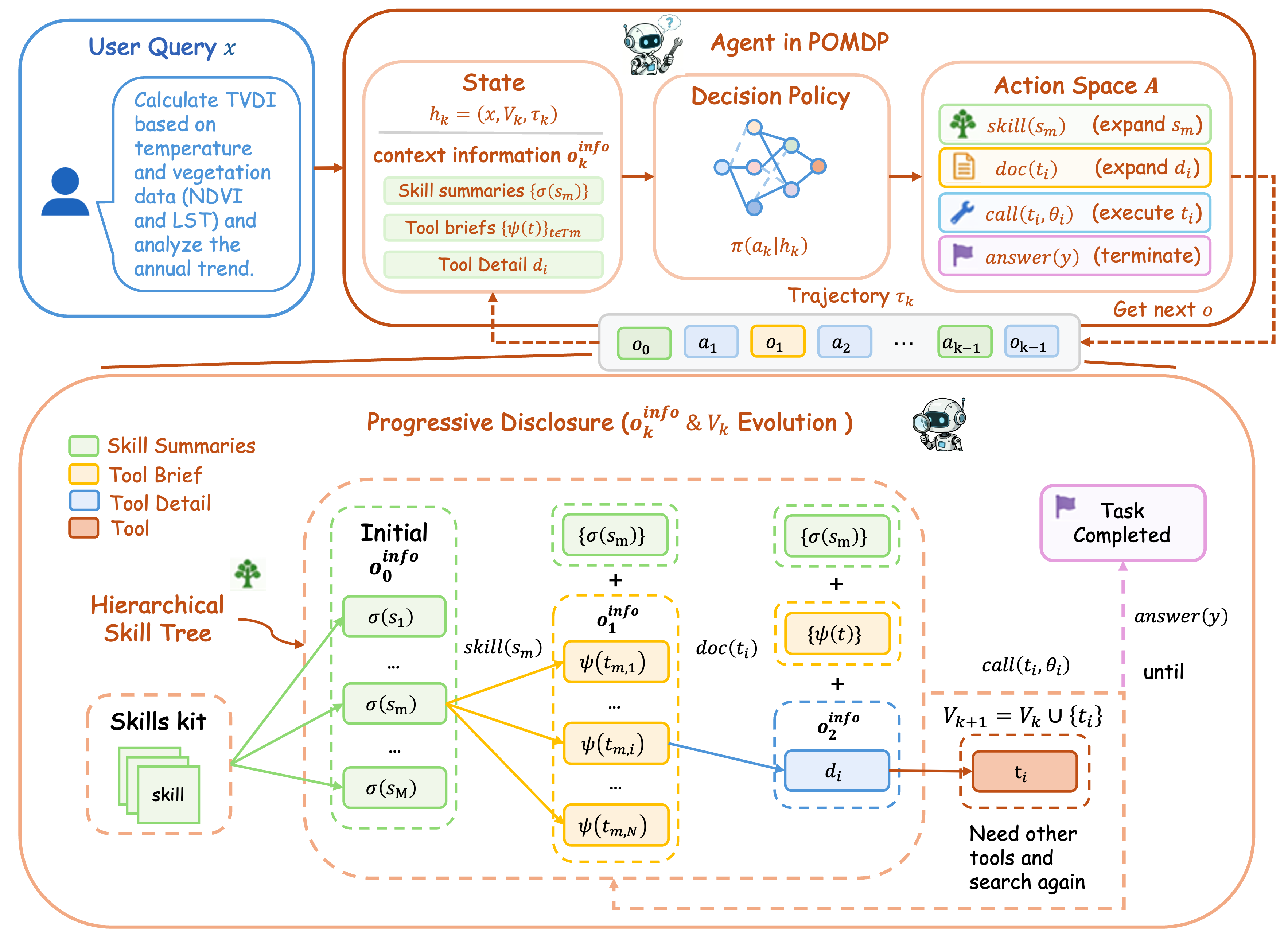}
\caption{The overall framework of the progressive active tool exploration mechanism based on a hierarchical skill tree. The top panel illustrates the unified sequential decision-making modeling, where the agent autonomously selects actions from the defined action space $\mathcal{A}$ based on the current decision state $h_k$. The bottom panel details the progressive disclosure strategy along the skill tree. Driven by the policy, the agent proactively executes exploration actions to sequentially acquire tool briefs and detailed execution documents, thereby progressively accumulating context observations $o_k^{\mathrm{info}}$ and dynamically expanding the callable tool set $\mathcal{V}_k$.}
\label{fig:framework}
\end{figure*}

This paper models the task-solving process of the RS agent as a sequential decision-making problem within a structured tool space, where information unfolding, tool discovery, tool execution, and result generation are uniformly integrated into a single policy space. Distinct from the traditional passive tool selection paradigm, this paper defines the agent as an active tool explorer. By explicitly modeling the dynamic evolution of the currently visible tool information set, the agent is empowered to autonomously determine when to expand the visible tool space and which specific subset of tool information to explore based on the current task, thereby advancing subsequent reasoning and execution. Building upon this core modeling, we further construct a hierarchical skill tree and design a corresponding progressive disclosure strategy. The overall methodology of RS-Claw comprises three main components: unified sequential decision-making modeling, hierarchical skill tree construction, and progressive disclosure strategy design, as illustrated in Fig.~\ref{fig:framework}.

\subsection{Unified Sequential Decision-Making Modeling}

Consider a user query $x$ and a large-scale RS tool library $\mathcal{T}=\{t_1,\dots,t_N\}$, where each tool $t_i$ corresponds to a semantic description $d_i$ used to describe its functionality, inputs, and outputs. Let $\mathcal{D}=\{d_1,\dots,d_N\}$ be the set of all full tool descriptions. RS tasks typically have long-horizon dependencies and heterogeneous intermediate representations, and their subsequent decisions depend on the intermediate observations generated by preceding steps. Therefore, the solving of a user query $x$ is essentially not a single-step matching problem, but a sequential decision-making problem that continuously evolves with intermediate observations.

Therefore, the task-solving process of the RS agent can be formulated as a Partially Observable Markov Decision Process (POMDP)~\cite{kaelbling1998planning}:
\begin{equation}
\mathcal{M}=(\mathcal{Z},\mathcal{A},\mathcal{O},P)
\end{equation}
where $\mathcal{Z}$ denotes the latent environmental state space, including the real RS data states and the environmental states generated after tool execution; $\mathcal{A}$ denotes the agent's action space; $\mathcal{O}$ denotes the observation space; $P(z_{k+1}\mid z_k,a_k)$ denotes the state transition probability. Let $a_k \in \mathcal{A}$ denote the specific action taken by the agent when making a decision at the $k$-th step, and $o_k \in \mathcal{O}$ denote the observation returned by the environment after the action is executed. Since the latent environmental state $z_k$ is invisible to the agent, the agent can only make decisions based on the execution history, the currently callable tool set, and the user query. To this end, we define the unified decision state of the agent at the $k$-th step as:
\begin{equation}
h_k=(x,\mathcal{V}_k,\tau_k)
\end{equation}
where $\tau_k=(o_0,a_1,o_1,a_2,o_2,\dots,a_{k-1},o_{k-1})$ represents the interaction history before the agent makes the decision at the $k$-th step. Specifically, we let $o_0$ denote the initial context available to the agent before the first decision, while subsequent observations record the results actually observed during the reasoning process. $\mathcal{V}_k \subseteq \mathcal{T}$ represents the set of tools currently directly callable by the agent at the $k$-th step. A tool can be called if and only if the semantic description $d_i$ of the tool $t_i$ exists in the current interaction history $\tau_k$. Therefore, the callable tool set is defined as:
\begin{equation}
\mathcal{V}_k = \{t_i \in \mathcal{T} \mid d_i \in \tau_k\}
\end{equation}

The agent selects action $a_k$ according to the policy $\pi(a_k\mid h_k)$. After the action $a_k$ is executed, the environment transitions to the next latent state and returns a new observation $o_k$.

Based on the unified model established above, existing tool acquisition mechanisms can be reformulated as specifically constrained forms within this sequential decision-making framework. Among them, the ``Flat'' strategy passively and statically injects the full tool descriptions $\mathcal{D}$ into the context. Its decision state can be written as:
\begin{equation}
h_k^{\mathrm{flat}}=(x,\mathcal{V}_k,\tau_k), \quad \text{where } o_0=\mathcal{D},\ \mathcal{V}_k=\mathcal{T}
\end{equation}
This formulation implies that throughout the entire decision-making process, all tools are defaulted to be directly callable, compelling the agent to make every decision under the interference of massive and voluminous tool descriptions. Consequently, its selectable actions at the $k$-th step can be written as:
\begin{equation}
a_k^{\mathrm{flat}}\in\{\mathrm{call}(t_i,\theta_i),\mathrm{answer}(y)\}, \quad t_i\in\mathcal{V}_k
\end{equation}
where $\mathrm{call}(t_i,\theta_i)$ denotes invoking the tool $t_i$ with parameters $\theta_i$ under the condition that $t_i$ belongs to the current callable set $\mathcal{V}_k$; $\mathrm{answer}(y)$ denotes outputting the final answer and terminating the reasoning. At this time, its observation information is solely manifested as the tool execution result $o_k^{\mathrm{exec}}$.

Similarly, the external retrieval paradigm based on the ``RAG'' strategy can also be viewed as a special case of the aforementioned decision-making process. This paradigm recalls a candidate subset from the full tool descriptions $\mathcal{D}$ via external algorithms, with the update rules falling completely outside the agent's action space. Its state also follows the unified format, but both the initial context and the callable tools are passively determined by the external retrieval results:
\begin{equation}
\begin{aligned}
h_k^{\mathrm{RAG}}&=(x,\mathcal{V}_k,\tau_k),\\
\text{where } o_0&=f_{\mathrm{sim}}(x,\mathcal{D}),\quad \mathcal{V}_k=\{t_i \mid d_i\in o_0\}
\end{aligned}
\end{equation}
where $f_{\mathrm{sim}}$ denotes the similarity retrieval function independent of the agent's action space. At this time, the initial context $o_0$ and the callable tool set $\mathcal{V}_k$ are both generated by the external retriever, and the agent cannot proactively expand this set through its own actions. Its available actions at the $k$-th step is the same as the flat paradigm:
\begin{equation}
a_k^{\mathrm{RAG}}\in\{\mathrm{call}(t_i,\theta_i), \mathrm{answer}(y)\}, \quad t_i\in\mathcal{V}_k
\end{equation}
Its observation information is likewise primarily manifested as the tool execution result $o_k^{\mathrm{exec}}$.

Through the above formulations, it becomes unequivocally clear that existing methods impose static constraints on the update of the callable set $\mathcal{V}_k$, forcibly isolating it from the agent's action space $\mathcal{A}$. Such constraints cause these methods to degenerate into a passive paradigm for tool selection. Under this passive paradigm, the agent acts merely as a receiver of a predefined toolset, incapable of autonomously guiding the information unfolding process.

Distinct from the aforementioned methods, this paper fundamentally internalizes the acquisition of tool information as an autonomous decision variable of the agent, proposing an active paradigm for tool exploration. The decision state of our proposed method can be formulated as:
\begin{equation}
h_k^{\mathrm{active}}=(x,\mathcal{V}_k,\tau_k), \quad \text{where } \mathcal{V}_k\subseteq\mathcal{T}
\end{equation}
Its optional action at the $k$-th step consists of information exploration actions, tool execution actions, and termination actions:
\begin{equation}
a_k^{\mathrm{active}} \in \mathcal{A}_{\mathrm{explore}} \cup \{\mathrm{call}(t_i,\theta_i), \mathrm{answer}(y)\}
\end{equation}
where $\mathcal{A}_{\mathrm{explore}}$ denotes the tool information acquisition actions that the agent can proactively perform. Its single-step observation output satisfies:
\begin{equation}
o_k =
\begin{cases}
o_k^{\mathrm{info}}, & a_k \in \mathcal{A}_{\mathrm{explore}},\\
o_k^{\mathrm{exec}}, & a_k = \mathrm{call}(t_i,\theta_i).
\end{cases}
\end{equation}
Here, the information observation $o_k^{\mathrm{info}}$ corresponds to the local tool information acquired by the agent during exploration actions, while the execution observation $o_k^{\mathrm{exec}}$ corresponds to the results of tool invocations. In stark contrast to the passive paradigm, the history trajectory $\tau_k$ in our method explicitly accumulates tool-related observations in tandem with the agent's active exploration, thereby subjecting the callable set $\mathcal{V}_k$ to the control of the agent's policy. Regarding how to impose a high-dimensional structured organization upon the massive tool library, as well as how to design $\mathcal{A}_{\mathrm{explore}}$ and the information observation $o_k^{\mathrm{info}}$ to support the progressive disclosure of tool information, we will elaborate on these in Sections~III-B and~III-C.

\begin{table*}[!t]
\caption{Comparison of mathematical modeling across different tool acquisition paradigms in the sequential decision-making process.\label{tab:paradigm_comparison}}
\centering
\begin{tabular}{p{3.0cm}p{3.3cm}p{3.8cm}p{4.3cm}}
\toprule
\multirow{2}{*}{\textbf{Comparison Dimension}} & \multicolumn{2}{c}{\textbf{Passive Paradigm}} & \textbf{Active Paradigm} \\
\cmidrule(lr){2-3}\cmidrule(lr){4-4}
& \textbf{Flat Paradigm} & \textbf{RAG Paradigm} & \textbf{Ours (Active Exploration)} \\
\midrule
Initial Context ($o_0$) & Full tool descriptions $\mathcal{D}$ & Subset recalled via external similarity function $f_{\mathrm{sim}}(x,\mathcal{D})$ & Minimal initial context (without underlying tool details) \\
Callable Tool Set ($\mathcal{V}_k$) & Statically global and invariant: $\mathcal{V}_k = \mathcal{T}$ & Statically local and invariant: $\mathcal{V}_k = \{t_i \mid d_i \in o_0\}$ & Autonomously and dynamically evolving with history trajectory: $\mathcal{V}_k \subseteq \mathcal{T}$ \\
Single-Step Action ($a_k$) & $a_k \in \{\mathrm{call}, \mathrm{answer}\}$ & $a_k \in \{\mathrm{call}, \mathrm{answer}\}$ & $a_k \in \mathcal{A}_{\mathrm{explore}} \cup \{\mathrm{call}, \mathrm{answer}\}$ \\
Single-Step Observation ($o_k$) & Only tool execution results $o_k^{\mathrm{exec}}$ & Only tool execution results $o_k^{\mathrm{exec}}$ & Incorporates local information observations $o_k^{\mathrm{info}}$ and execution results $o_k^{\mathrm{exec}}$ \\
Tool Acquisition Mechanism & Forced one-time global context registration & Passive recall independent of the agent's actions & Autonomous search internalized within the agent \\
Probabilistic Representation of Acquisition & Global conditional distribution relying on full static documents & Truncated distribution based on external independent vector similarity & Autoregressive conditional probability based on previous context tokens \\
\bottomrule
\end{tabular}
\end{table*}

\subsection{Hierarchical Skill Tree Construction}

As modeled in Section~III-A, the dynamic acquisition of tool information is abstracted as the joint evolution of the history trajectory $\tau_k$ and the callable set $\mathcal{V}_k$. However, under the traditional flat organizational structure, when facing the full tool descriptions $\mathcal{D}$, the agent lacks explicit search boundaries and effective observation targets. To address this, inspired by the top-down hierarchical cognitive logic of humans---from macroscopic intentions to microscopic actions---and hierarchical abstraction in sequential decision-making~\cite{sutton1999between}, this paper introduces a hierarchical skill tree topological structure. This imposes an organizational framework upon the high-dimensional tool space and provides priors for subsequent local information unfolding. Specifically, let $\mathcal{S}=\{s_1,\dots,s_M\}$ be the set of skill nodes. We perform an orthogonal partition on the full tool set $\mathcal{T}$, mapping it precisely into these $M$ skill nodes to form mutually exclusive local tool subspaces $\mathcal{T}_m$. This strict constraint ensures that any underlying RS tool $t_i \in \mathcal{T}$ uniquely belongs to a single parent skill node $s_m$. Building upon this, we rigorously decouple the massive tool space into three logically progressive information subspaces, which jointly constitute the complete hierarchical skill tree space $\mathcal{I}$ to guide the agent in top-down sequential exploration:

\begin{enumerate}
\item \textbf{Skill Summary Layer:} This layer consists of $M$ high-dimensional skill node summaries $\sigma(s_m)$, primarily serving to provide a macroscopic functional index for the entire toolset. Specifically, we employ a concise natural language segment as $\sigma(s_m)$ to demarcate the core functional boundary of the branch for the agent, thereby guiding the large language model to rapidly accomplish semantic matching between the macroscopic task intention and the local tool cluster. This global capability view, constructed at an extremely low token cost, serves directly as the initial context $o_0$ for the agent during actual reasoning. It assists the agent in conducting a coarse screening of macroscopic directions in the early decision-making stages, effectively preventing it from deviating into completely irrelevant domain branches, and rapidly narrowing the search scope down to a specific capability subspace.

\item \textbf{Tool Catalog Layer:} Once the search scope is narrowed, the agent encounters multiple candidate tools within a specific branch. This layer facilitates local disambiguation by providing brief descriptions $\psi(t_i)$ of these tools. Given that tools within a specific skill subspace often exhibit high functional similarity, this layer deliberately obscures underlying code-level parameter specifications to guarantee decision precision, retaining only core functional descriptions, applicability boundaries, and high-level input/output dependencies. This filtering mechanism enables the agent to compare similar tools free from the interference of underlying implementation details. Only after the target tool is definitively selected does the decision process advance to the next stage.

\item \textbf{Tool Documentation Layer:} Functioning as the interface for the agent to interact with the environment, this layer is situated at the terminal leaves of the tree topology. It encompasses the complete semantic descriptions $d_i$ of the underlying tools, covering detailed execution documents, strict API signatures, and parameter specifications to support the ultimate tool invocation actions. These machine-readable execution specifications typically contain numerous cumbersome constraints, which not only cause token overhead surges but also easily induce attention defocus in LLMs over long contexts. Implementing lazy loading for these detailed pieces of information ensures that the agent only loads and bears this information payload on-demand after fully clarifying its invocation intention, thereby maximizing the stability of long-horizon reasoning and the legality of tool invocations.
\end{enumerate}

This three-layer architecture transcends a mere segmentation of original tools to represent a profound alignment with the RS task-solving logic. Relying on the structured organization between adjacent layers, this architecture strictly adheres to the progressive search path of first determining the capability scope, then screening specific tools, and finally loading execution parameters, enabling the originally unstructured, flat tool set to evolve into a well-structured tree topology. During the agent's top-down, layer-by-layer reasoning process, the exploration granularity is continuously refined, while irrelevant semantic noise is physically isolated and progressively filtered out, thus maintaining a pure context environment for the large language model when executing long-horizon tasks.

\subsection{Progressive Disclosure Strategy Design}

Based on the static hierarchical skill tree space $\mathcal{I}$ constructed in Section~III-B, this section focuses on designing the agent's dynamic exploration mechanism over this space, directly instantiating the unfolding logic of local tool information into the agent's policy itself. Initially, the system only provides the topmost skill layer summaries to the agent as the initial context, and no underlying tools can be directly called yet, i.e.:
\begin{equation}
o_0=\{\sigma(s_m)\}_{m=1}^{M}, \qquad \mathcal{V}_1=\emptyset
\end{equation}

At the $k$-th step, the agent selects an action based on the current decision state $h_k$. Following the unified definition in Section~III-A, the information exploration actions $\mathcal{A}_{\mathrm{explore}}$ in our method specifically comprise two categories: one is $\mathrm{skill}(s_m)$, which expands the brief descriptions $\psi(t)$ of all candidate tools under the skill node $s_m$; the other is $\mathrm{doc}(t_i)$, which expands the detailed document $d_i$ of the tool $t_i$. The observations returned by the information exploration actions will simultaneously affect the history trajectory and the subsequent tool calling permissions.

When the agent executes $a_k=\mathrm{skill}(s_m)$, the system only returns the brief descriptions of all tools under that branch, $o_k^{\mathrm{info}}=\{\psi(t)\mid t\in\mathcal{T}_m\}$. The role of this observation $o_k^{\mathrm{info}}$ is to help the agent narrow down the search scope to precisely locate the required tool, but it will not directly unlock the calling permission for the underlying tools (i.e., $\mathcal{V}_{k+1}=\mathcal{V}_k$). Only when the agent further executes $a_k=\mathrm{doc}(t_i)$ will the system return the detailed execution document $d_i$ of the tool $t_i$. At this time, not only will this document be written into the history trajectory, but the underlying tool $t_i$ will also be formally incorporated into the callable set in the next step (i.e., $\mathcal{V}_{k+1}=\mathcal{V}_k \cup \{t_i\}$). Furthermore, for regular execution actions $a_k=\mathrm{call}(t_i,\theta_i)$ or the termination action $a_k=\mathrm{answer}(y)$, the environment solely returns the physical execution results or a termination flag, without introducing any new available tools (i.e., $\mathcal{V}_{k+1}=\mathcal{V}_k$).

This update rule indicates that the skill layer and the tool catalog layer return intermediate information observations, which are written into the history trajectory and influence subsequent reasoning, but do not immediately make the underlying tools callable. Only after the agent proactively requests the detailed document of a specific tool is that document $d_i$ incorporated into the history trajectory, and the corresponding tool $t_i$ enters the callable set $\mathcal{V}_{k+1}$. Therefore, tool discovery is no longer manifested as a one-time statically injected context, but rather a sequential process autonomously advanced by the agent.

Because only local information relevant to the current task is unfolded at each step, our method compresses the context load from a global $\mathcal{O}(N)$ information overhead down to a local $\mathcal{O}(K)$ information overhead, where $K \ll N$ and is dynamically determined by the policy. More importantly, this mechanism can reduce the influx of irrelevant information into the context, thereby alleviating context pollution and token consumption caused by full-set registration. Consequently, the abstract active tool exploration process proposed in Section~III-A is concretely implemented at runtime as a policy that progressively accumulates tool-related observations along the hierarchical skill tree and synchronously expands the callable tool set $\mathcal{V}_k$.

\section{Experiments}

\subsection{Experimental Setup}

\textbf{Benchmark.} We evaluate RS-Claw on the Earth-Bench, which comprises 248 RS agent analysis questions spanning multiple RS tasks, including spectral index computation, time-series analysis, change detection, building and object detection, and multi-source data fusion. The benchmark defines two evaluation modes: Autonomous Planning mode (AP), corresponding to implicit step specification, which assesses the agent's ability to autonomously plan solution paths---the agent must independently decide which tools to invoke and in what order; and Instruction Following mode (IF), corresponding to explicit step specification, which assesses the agent's ability to translate human instructions into executable actions---reasoning steps are provided externally, and the agent is responsible for converting them into correct tool calls. Since the official results were obtained in an environment where the ChangeOS dependency was available, whereas the released benchmark does not provide the required deep learning model or precomputed outputs for this tool, we exclude 14 ChangeOS-related questions (IDs 216--225 and 242--245) from all reported results for a fair comparison; consequently, all metrics are computed over 234 questions.

\textbf{Baselines.} We compare RS-Claw, which actively explores the tool space, against two passive-selection baseline agents:
\begin{enumerate}
\item Flat, the official Earth-Agent baseline, which writes all available tools as an unstructured flat list into the system prompt, from which the agent directly selects tools; and
\item RAG, a retrieval-augmented tool-aware agent that, before each question, retrieves the 19 most relevant tools from the tool library based on question semantics (with the file enumeration tool \texttt{get\_filelist} additionally forced-included), and provides the retrieval results as tool context for the agent's planning and invocation.
\end{enumerate}
Both baselines and RS-Claw are built on the ReAct framework, where agents complete multi-step tool-call reasoning through alternating Thought--Action--Observation cycles.

\textbf{Models.} To assess the generalizability of our method across different model capability levels, we conduct experiments on three large language models: GPT-5, DeepSeek-V3.1, and Qwen3-32b. Ablation studies are conducted exclusively on Qwen3-32b to control experimental cost.

\textbf{Skill tree configuration.} RS-Claw organizes tools into five skill categories consistent with the Earth-Agent official tool taxonomy: Index, Inversion, Perception, Analysis, and Statistics. Full tool listings and counts for each skill are provided in Appendix~\ref{app:skill_tree}.

\textbf{Evaluation metrics.} We evaluate each method along two dimensions: accuracy and context overhead. For accuracy, we adopt the two-tier evaluation protocol of the Earth-Bench. End-to-end metrics include Accuracy and Efficiency; step-level metrics include Tool-Any-Order, Tool-In-Order, Tool-Exact-Match, and Parameters. The main tables report the three core metrics---Accuracy, Tool-Any-Order, and Tool-In-Order---with complete results provided in Appendix~\ref{app:full_results}. For context overhead, we measure the impact of different tool disclosure strategies on context length using the average number of input tokens per question and per turn.

\subsection{Experimental Results}

\begin{table*}[!t]
\caption{Accuracy metrics comparison. Accuracy metrics of RS-Claw (active tool exploration) against two passive tool selection baselines (Flat, RAG) across three models and two evaluation modes (AP and IF). Complete metrics are provided in Appendix Table~\ref{tab:full_accuracy}.\label{tab:accuracy}}
\centering
\begin{tabular}{llcccccc}
\toprule
\multirow{2}{*}{Model} & \multirow{2}{*}{Method} & \multicolumn{2}{c}{Accuracy} & \multicolumn{2}{c}{Tool-Any-Order} & \multicolumn{2}{c}{Tool-In-Order} \\
\cmidrule(lr){3-4}\cmidrule(lr){5-6}\cmidrule(lr){7-8}
& & AP & IF & AP & IF & AP & IF \\
\midrule
\multirow{3}{*}{GPT-5} & Flat & 65.67 & 64.38 & 67.40 & 69.86 & 55.89 & 58.74 \\
& RAG & 59.23 & 60.09 & 58.63 & 59.05 & 49.66 & 49.90 \\
& RS-Claw (Ours) & \textbf{68.67} & \textbf{70.82} & \textbf{73.10} & \textbf{75.57} & \textbf{59.52} & \textbf{62.63} \\
\midrule
\multirow{3}{*}{DeepSeek-V3.1} & Flat & 49.36 & 51.07 & 77.25 & 76.66 & 61.19 & 62.19 \\
& RAG & 39.91 & 46.78 & 62.39 & 61.22 & 47.82 & 49.97 \\
& RS-Claw (Ours) & \textbf{57.08} & \textbf{55.79} & \textbf{80.77} & \textbf{81.17} & \textbf{68.62} & \textbf{69.44} \\
\midrule
\multirow{3}{*}{Qwen3-32b} & Flat & 20.60 & 25.32 & 39.86 & 42.36 & 22.14 & 34.31 \\
& RAG & 20.17 & 22.75 & 30.04 & 35.36 & 20.75 & 26.06 \\
& RS-Claw (Ours) & \textbf{33.05} & \textbf{29.18} & \textbf{50.22} & \textbf{60.77} & \textbf{32.06} & \textbf{44.91} \\
\bottomrule
\end{tabular}
\end{table*}

\begin{table*}[!t]
\caption{Context overhead metrics comparison. Average input token counts of RS-Claw (active tool exploration) against two passive tool selection baselines (Flat, RAG) across three models and two evaluation modes (AP and IF), measured in tokens/question and tokens/turn.\label{tab:context}}
\centering
\begin{tabular}{llcccc}
\toprule
\multirow{2}{*}{Model} & \multirow{2}{*}{Method} & \multicolumn{2}{c}{Tokens per question} & \multicolumn{2}{c}{Tokens per turn} \\
\cmidrule(lr){3-4}\cmidrule(lr){5-6}
& & AP & IF & AP & IF \\
\midrule
\multirow{3}{*}{GPT-5} & Flat & 134,076 & 129,499 & 21,833 & 21,754 \\
& RAG & \textbf{39,080} & \textbf{37,280} & \textbf{7,394} & \textbf{7,155} \\
& RS-Claw (Ours) & 107,428 & 100,302 & 7,945 & 7,708 \\
\midrule
\multirow{3}{*}{DeepSeek-V3.1} & Flat & 352,398 & 289,428 & 27,587 & 27,490 \\
& RAG & \textbf{111,566} & \textbf{97,659} & 9,147 & 9,143 \\
& RS-Claw (Ours) & 158,433 & 159,347 & \textbf{8,116} & \textbf{8,199} \\
\midrule
\multirow{3}{*}{Qwen3-32b} & Flat & 502,119 & 517,445 & 30,612 & 31,756 \\
& RAG & 105,383 & 116,778 & 8,999 & 9,579 \\
& RS-Claw (Ours) & \textbf{70,759} & \textbf{97,325} & \textbf{5,951} & \textbf{6,309} \\
\bottomrule
\end{tabular}
\end{table*}

\textbf{Accuracy analysis.} As shown in Table~\ref{tab:accuracy}, RS-Claw achieves higher accuracy than both passive baselines across all three models and both evaluation modes, with tool-matching metrics also leading comprehensively, validating the effectiveness of the progressive skill tree disclosure strategy. Notably, the improvement margin increases as model capability decreases: in AP mode on Qwen3-32b, RS-Claw outperforms Flat by 12.45 percentage points, far exceeding the 3.00-point gain on GPT-5. This indicates that progressive disclosure effectively alleviates constrained reasoning space and attention defocus across models of varying capability levels, with models more sensitive to context load demonstrating larger gains. RAG achieves lower accuracy than RS-Claw on all three models, with weaker step-level metrics as well. As a passive tool recipient, the RAG agent delegates tool selection to an external embedding model rather than the agent itself; the LLM cannot actively adjust the tool scope based on intermediate results emerging during reasoning, making it difficult to cover the critical tools required by subsequent reasoning steps in long-horizon tasks. This constitutes an inherent limitation of insufficient tool coverage in multi-step RS analysis tasks. The two passive baselines thus fail on opposite ends of the trade-off---context overload versus incomplete tool coverage---while RS-Claw's active exploration mechanism achieves an effective balance on both dimensions.

\textbf{Context overhead analysis.} As shown in Table~\ref{tab:context}, RS-Claw substantially reduces the total input tokens per question compared to Flat---in AP mode on Qwen3-32b, the input token compression ratio reaches approximately 86\%, with tokens per turn reduced by 81\%, directly validating the core design objective of progressive disclosure: on-demand tool loading that dramatically reduces context overhead. Consistent with the accuracy analysis, the compression ratio also varies with model capability: GPT-5 achieves a compression ratio of approximately 20\%, far lower than that of Qwen3-32b, indicating that the context savings from progressive disclosure differ across models, with models more sensitive to context load benefiting more substantially. RS-Claw incurs lower tokens per turn than RAG overall, owing to the fact that RS-Claw loads only the currently relevant nodes per turn, whereas RAG carries a fixed number of tool descriptions every turn. Notably, the total per-question cost advantage of RS-Claw over RAG varies across models: on Qwen3-32b, RS-Claw incurs lower total overhead than RAG, whereas on GPT-5, RS-Claw incurs higher total overhead than RAG---a discrepancy attributable to differences in the depth of skill tree exploration across models. Taken together, the two passive baselines occupy opposite ends of the context-overhead spectrum---Flat trades efficiency for completeness, while RAG trades tool coverage for lower overhead---whereas RS-Claw's active exploration mechanism keeps context consumption locally bounded while preserving tool coverage, achieving an effective balance between the two.

\subsection{Ablation Study}

In this section, we validate the necessity of two core design dimensions in RS-Claw's progressive disclosure strategy through two sets of controlled variants: the semantic organizational structure of the skill tree, and the progressive disclosure mechanism that retains the skill summary layer within the three-tier skill tree. We compare the following three variants on Qwen3-32b: RS-Claw (the agent used in the main experiments, employing semantically grouped three-tier skill tree and two-step progressive disclosure); Random (retaining the three-tier skill tree and two-step progressive disclosure but randomly assigning tools to five skill nodes, disrupting the semantic coherence of the skill summary layer); and 2layers (removing the skill summary layer, retaining only the tool catalog layer and tool documentation layer, and embedding all tool names and brief descriptions by category directly into the system prompt, from which the agent selects target tools and then calls doc to retrieve detailed execution documents---degenerating to a one-step disclosure containing only the doc information exploration action without the skill action. This represents an intermediate state between the flat Flat paradigm and RS-Claw). Three variants form an ordered gradient in tool disclosure structure: Flat injects all tool descriptions into the context at once; 2layers retains full tool-name visibility but loads detailed execution documents on demand, essentially degenerating the three-tier skill tree of Section~III-B to two tiers; RS-Claw fully internalizes the progressive unfolding of tool information into the agent's sequential decision-making through the complete three-tier structure. Complete accuracy metrics are provided in Appendix Table~\ref{tab:ablation_full}.

\begin{table}[!t]
\caption{Ablation accuracy metrics for progressive disclosure strategy. Accuracy metrics of the two ablation variants (Random, 2layers) compared with RS-Claw under AP and IF modes.\label{tab:ablation_acc}}
\centering
\begin{tabular}{lcccccc}
\toprule
\multirow{2}{*}{Method} & \multicolumn{2}{c}{Accuracy} & \multicolumn{2}{c}{Tool-Any-Order} & \multicolumn{2}{c}{Tool-In-Order} \\
\cmidrule(lr){2-3}\cmidrule(lr){4-5}\cmidrule(lr){6-7}
& AP & IF & AP & IF & AP & IF \\
\midrule
Random & 23.18 & 25.32 & 44.21 & 48.13 & 31.44 & 37.93 \\
2layers & 25.75 & 25.32 & \textbf{58.07} & \textbf{62.50} & \textbf{43.16} & \textbf{50.37} \\
RS-Claw (Ours) & \textbf{33.05} & \textbf{29.18} & 50.22 & 60.77 & 32.06 & 44.91 \\
\bottomrule
\end{tabular}
\end{table}

\begin{table}[!t]
\caption{Ablation context overhead for progressive disclosure strategy. Average input token counts of the two ablation variants (Random, 2layers) compared with RS-Claw under AP and IF modes.\label{tab:ablation_ctx}}
\centering
\begin{tabular}{lcccc}
\toprule
\multirow{2}{*}{Method} & \multicolumn{2}{c}{Tokens per question} & \multicolumn{2}{c}{Tokens per turn} \\
\cmidrule(lr){2-3}\cmidrule(lr){4-5}
& AP & IF & AP & IF \\
\midrule
Random & 100,903 & 139,269 & 6,337 & 7,525 \\
2layers & 82,896 & 106,165 & 6,415 & 7,196 \\
RS-Claw (Ours) & \textbf{70,759} & \textbf{97,325} & \textbf{5,951} & \textbf{6,309} \\
\bottomrule
\end{tabular}
\end{table}

\textbf{Necessity of semantic organizational structure of the skill tree.} As shown in Tables~\ref{tab:ablation_acc} and~\ref{tab:ablation_ctx}, Random achieves 9.87 percentage points lower accuracy than RS-Claw in AP mode, while tokens per question increase by 43\%. When skill groupings are randomized, the semantic prior provided by the skill layer becomes invalid; the agent can no longer effectively narrow the candidate set based on skill summaries and must explore incorrect skill nodes multiple times to locate target tools. These futile exploration turns accumulate continuously in the context, exacerbating restricted reasoning space and attention defocus, ultimately degrading both accuracy and context overhead.

\textbf{Necessity of retaining the skill summary layer in the progressive disclosure mechanism.} As shown in Tables~\ref{tab:ablation_acc} and~\ref{tab:ablation_ctx}, 2layers exposes all tool names directly in the system prompt, bypassing the skill information exploration action required to route to the skill summary layer for tool selection. Its step-level metrics outperform RS-Claw, as the improved tool visibility directly enhances tool discovery. However, under Qwen3-32b, this gain comes at the cost of reasoning space---a cost directly reflected in the contradiction where 2layers achieves a higher Tool-Any-Order (58.07) than RS-Claw (50.22) yet a lower Accuracy (25.75) versus RS-Claw (33.05): the improvement in tool discovery does not translate into higher final answer accuracy, indicating that the compression of reasoning space offsets the benefit of improved tool visibility, ultimately yielding lower accuracy. This directly validates the judgment of Section~III-B from the perspective of hierarchical skill tree design: the skill summary layer is the key barrier controlling context scale---omitting it causes the two-tier structure to surpass the three-tier in tool discovery yet fall significantly behind in end-to-end accuracy. Appendix~\ref{app:case_2layers} provides two concrete trajectory comparisons that further illustrate this mechanism: both pairs share identical Tool-Any-Order and Tool-In-Order scores, ruling out any difference in tool discovery ability and attributing failure directly to constrained reasoning space after omitting the skill summary layer, manifesting as intermediate-file confusion and premature truncation of multi-step planning.

Together, the two ablation groups demonstrate that the semantic organizational structure of the skill tree and the progressive disclosure mechanism retaining the skill summary layer are both indispensable. The former provides the agent with an effective prior for selecting the skill summary layer; the latter ensures that tool information is expanded on demand. These two mechanisms jointly constrain the tool information entering the context to the locally relevant subset for the current task, thereby avoiding constrained reasoning space.

\subsection{Tool Scalability Experiments}

This section examines the robustness of the progressive disclosure strategy under continuous expansion of the tool library along two dimensions: same-domain scaling (starting from the minimum tool set required for the task, progressively adding same-domain irrelevant tools up to the full set of 104) and cross-domain scaling (injecting cross-domain tools entirely unrelated to RS tasks into the tool library). All experiments use the Qwen3-32b model.

\subsubsection{Same-Domain Tool Scaling}

The experiment fixes the skill package structure and, starting from GT (the minimum tool set required to complete tasks, extracted from the Earth-Bench official ground-truth trajectories), progressively injects same-domain irrelevant tools into the tool library in increments of 20 until all 104 tools are covered (All Tools, i.e., the RS-Claw configuration in the main experiments), forming six scale gradients. The goal is to examine how context overhead and task accuracy respond to increasing total tool count under the two disclosure paradigms. Complete numerical results are provided in Appendix Tables~\ref{tab:scaling_same_acc} and~\ref{tab:scaling_same_ctx}.

\begin{figure}[!t]
\centering
\includegraphics[width=\columnwidth]{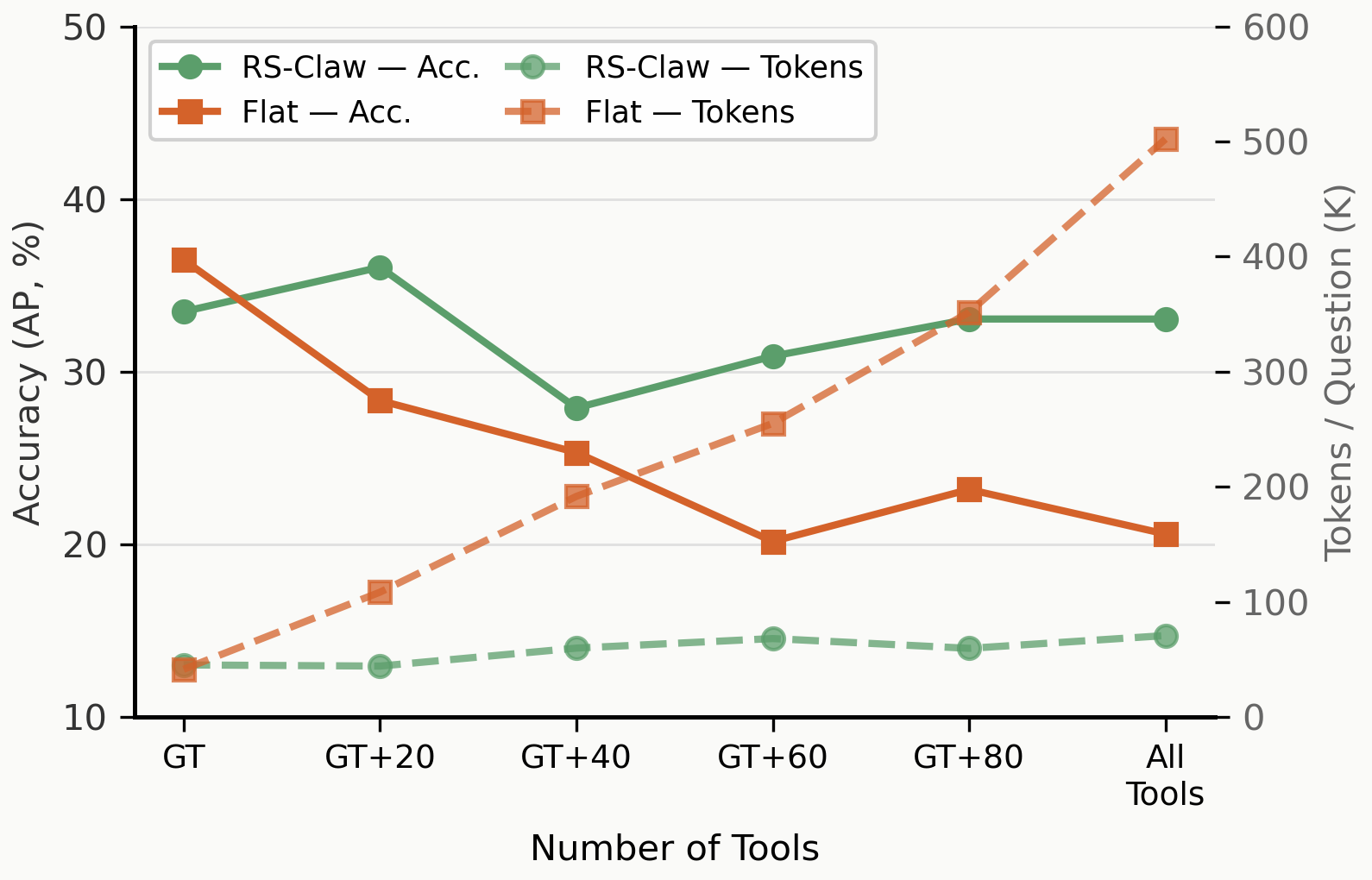}
\caption{Accuracy and context overhead curves under same-domain tool scaling. AP mode accuracy (left axis) and per-question token overhead (right axis, K) of RS-Claw and Flat as the tool library scales from the minimum set (GT) to the full toolset (104 tools).}
\label{fig:scaling_same}
\end{figure}

As shown in Fig.~\ref{fig:scaling_same}, when the tool library contains only the minimum required tool set (GT), Flat achieves slightly higher accuracy and slightly lower per-question token overhead than RS-Claw---under zero-redundancy conditions, full registration introduces no semantic noise, and direct tool visibility eliminates the extra navigation overhead of the skill layer, allowing Flat's direct-visibility advantage to fully manifest. This comparison indicates that the advantage of RS-Claw over Flat is not unconditional, but rather stems from the context load introduced by tool scale expansion. As tool count increases, the gap rapidly reverses: RS-Claw accuracy remains broadly stable while Flat declines continuously, falling below RS-Claw after GT+20 with the gap generally widening thereafter; token overhead likewise diverges markedly, with RS-Claw growing gradually while Flat expands near-linearly, reaching a growth of over 1,100\% in AP mode. A similar trend is observed in IF mode. By constraining visible tool information to a locally bounded scope at all times, progressive disclosure suppresses both context inflation and the performance degradation caused by escalating candidate interference as tool scale grows, whereas Flat deteriorates continuously on both accuracy and overhead dimensions as tool count increases.

\subsubsection{Cross-Domain Tool Scaling}

Building on the same-domain scaling experiments, this section further examines the behavior of the two disclosure paradigms when the tool library is injected with cross-domain tools entirely unrelated to RS tasks. Starting from the Earth-Bench (104 RS tools), we sequentially inject API-Bank~\cite{li2023api} (75 general-purpose tools covering account authentication, calendar reminders, financial services, healthcare, etc., totaling 179 tools, denoted stage1) and ToolBench~\cite{qin2023toolllm} (55 tools covering advertising, business, music, etc., further expanding to 234 tools on top of stage1, denoted stage2). All newly added tools provide no practical utility for Earth-Bench questions, constituting pure cross-domain semantic noise. Complete numerical results are provided in Appendix Tables~\ref{tab:scaling_cross_acc} and~\ref{tab:scaling_cross_ctx}.

\begin{figure}[!t]
\centering
\includegraphics[width=\columnwidth]{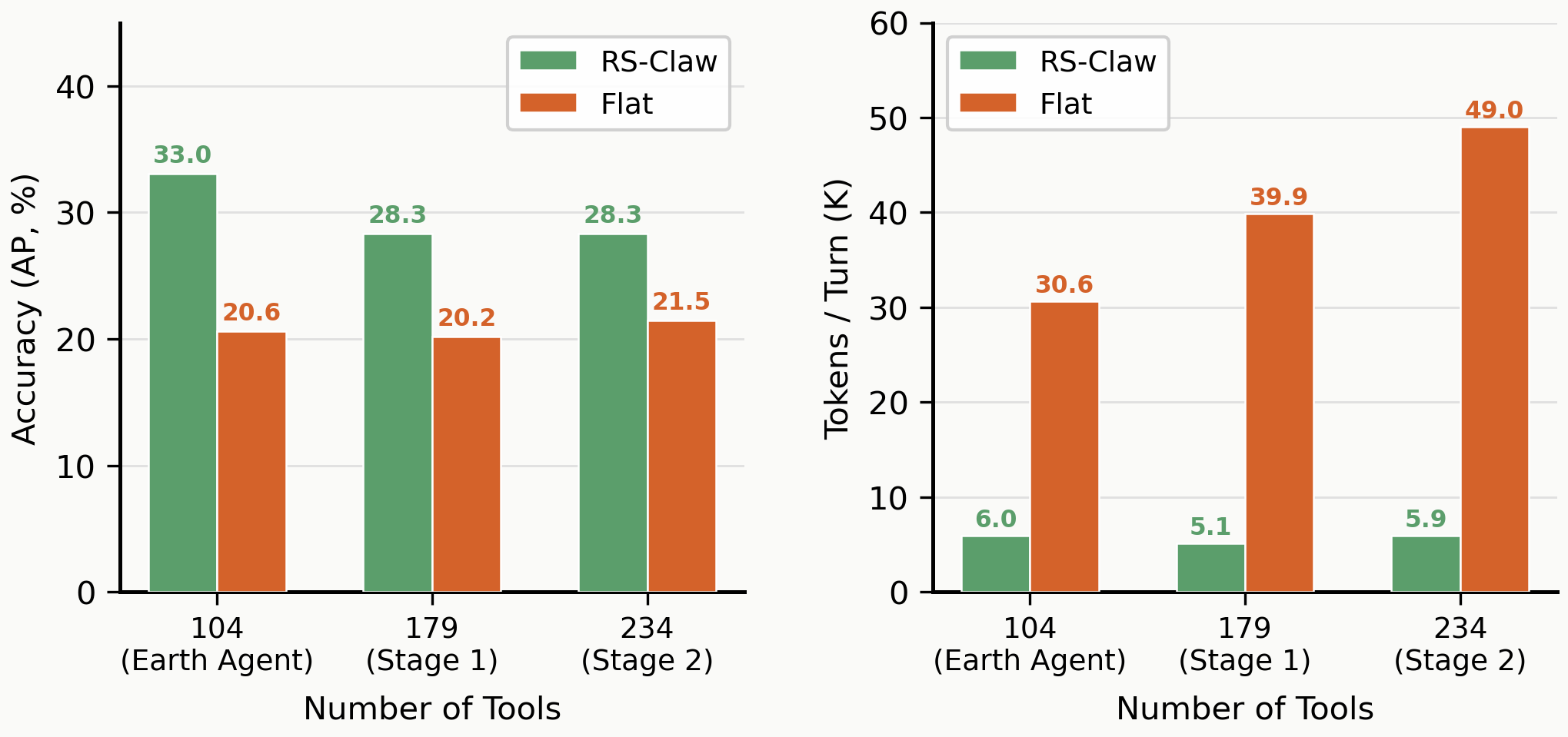}
\caption{Accuracy and context overhead comparison under cross-domain tool scaling. AP mode accuracy (left) and per-turn token overhead (right, K) of RS-Claw and Flat across three stages of cross-domain tool library expansion (104 $\to$ 179 $\to$ 234 tools).}
\label{fig:scaling_cross}
\end{figure}

As shown in Fig.~\ref{fig:scaling_cross}, after injecting cross-domain tools, neither paradigm experiences significant accuracy collapse---RS-Claw decreases modestly but remains broadly stable, while Flat consistently remains at a lower level. This differs from the continuous accuracy decline of Flat observed in the same-domain scaling experiments; a plausible explanation is that cross-domain tools are semantically distant from RS tasks, making it easier for the model to filter them out during reasoning. However, on the overhead dimension, the divergence remains pronounced: Flat tokens per turn grow linearly with total tool count, while RS-Claw remains essentially stable. This result directly validates the core advantage of progressive disclosure: the on-demand loading mechanism ensures that descriptions of irrelevant tools never enter the context, so tool library expansion has minimal impact on actual context consumption. Progressive disclosure thus combines controllable overhead with robust accuracy in open-world tool environments.

Taken together, the same-domain and cross-domain scaling experiments demonstrate that the progressive disclosure strategy exhibits favorable robustness under both typical tool library expansion scenarios: accuracy degradation is relatively gradual, and context overhead growth is substantially lower than that of Flat. This property endows the approach with stronger scalability for real-world applications characterized by dynamic growth of open-world tool libraries.

\section{Discussion}

\subsection{Limitations}

Although RS-Claw demonstrates consistent improvements over both baselines across multiple models and evaluation modes, several limitations remain.

First, the five skill categories in the current skill tree are directly inherited from the Earth-Agent official tool taxonomy, constituting a static, manually predefined topology. When the tool library undergoes substantial changes or is migrated to a new domain, domain experts are required to redesign the tree structure, and the existing grouping exhibits granularity imbalance across skill nodes.

Second, progressive disclosure incorporates tool exploration into the agent's decision space, yet the quality of exploration paths fundamentally depends on the planning capability of the underlying LLM; when the model's own planning ability is insufficient, the agent may still make suboptimal exploration decisions.

Third, the skill $\to$ doc two-step exploration introduces additional interaction rounds, and the advantage of progressive disclosure stems from the context load imposed by tool scale---in scenarios with a small-scale tool library, the extra exploration overhead may outweigh the benefits.

Finally, this work is validated solely on the Earth-Bench with only three models; the generalization of the method to more domains and larger-scale tool libraries warrants further investigation.

\subsection{Connections with other work}

\subsubsection{Relationship to existing RS agent systems}
Works such as RS-Agent, Earth-Agent, ChangeAgent, and GeoLLM-Squad have advanced RS tasks from single-step perception toward the automated execution of complex workflows. However, these systems share a common implicit assumption: the tool space is fully visible before task execution begins, and the agent plans and invokes tools over a static, predefined toolset. As the scale of specialized RS tool libraries continues to grow, the context load and declining tool selection efficiency arising from this assumption become increasingly pronounced. Building upon the ReAct reasoning framework established by these works, RS-Claw further incorporates tool acquisition itself into the agent's decision space, allowing the callable tool set to evolve dynamically throughout the reasoning process. This provides a more scalable tool management mechanism for long-horizon RS tasks without requiring any modification to the underlying model.

\subsubsection{Relationship to general-purpose tool selection and organization methods}
Whether retrieval-based methods such as ToolReAGt, Gorilla, and ToolLLM, or hierarchical organization methods such as AnyTool, ToolNet, and Graph of Skills, all share a common limitation: the triggering logic of tool acquisition is controlled by an external module and remains decoupled from the agent's reasoning process. The former relies on an external retriever to lock in the candidate toolset in a single pass before task execution; the latter introduces hierarchical structures, but the unfolding of those hierarchies is likewise driven by preset program logic rather than the agent's reasoning intent. The fundamental distinction of RS-Claw is that tool information acquisition is fully internalized as the agent's own decision variable. The exploration path is an output of the policy, and the callable tool set evolves continuously with intermediate reasoning states, enabling the agent to dynamically adjust its tool exploration direction according to actual task demands during long-horizon tasks.

\subsection{Significances}

\subsubsection{Significance for the RS domain}
RS tasks are characterized by long-horizon workflow chains, high semantic similarity among tool descriptions, and complex intermediate states --- precisely the conditions under which the full tool registration paradigm faces its most severe bottleneck. RS-Claw is designed directly against these domain-specific pain points, validating the effectiveness of the active exploration mechanism within multi-source RS tool spaces and offering a viable path toward automated execution of long-horizon RS tasks over large-scale specialized tool libraries.

\subsubsection{Practical significance}
RS-Claw operates entirely at the tool-side architecture level and requires no model fine-tuning, making it a plug-and-play context management strategy that can be directly integrated into any ReAct-based agent system. The method achieves substantial reductions in context overhead while maintaining accuracy gains over both baselines, demonstrating that lowering inference cost and preserving task completion quality are not mutually exclusive. Tool scaling experiments further show that RS-Claw's context overhead remains approximately stable as the tool library grows, a property that translates directly into engineering value for real-world deployment scenarios characterized by dynamically expanding tool libraries.

\subsubsection{Theoretical significance}
This paper unifies passive tool exploration paradigms (Flat and RAG) and the active exploration paradigm within a single sequential decision-making framework, casting both paradigm classes as special cases of the same formalization and providing a unified modeling perspective on the tool selection problem. The framework explicitly distinguishes between tool information acquisition and tool execution as two distinct types of observations, revealing that tool acquisition is inherently a dynamic process that evolves with the reasoning state rather than a one-shot static decision.

\section{Conclusion and Future Work}

\subsection{Conclusion}

This paper addresses the context bottleneck faced by RS agents operating over large-scale tool libraries, and proposes RS-Claw, a progressive active tool exploration mechanism based on a hierarchical skill tree. We internalize tool information acquisition as the agent's autonomous decision variable, incorporating tool exploration actions into the policy space through unified POMDP-based sequential decision-making modeling. A three-tier hierarchical skill tree coupled with a progressive disclosure strategy enables on-demand tool loading, effectively compressing tool context overhead. Experiments on the Earth-Bench demonstrate that RS-Claw outperforms both passive baselines (Flat and RAG) across three models and two evaluation modes, with improvement margins increasing as model capability decreases and context compression reaching up to 86\%. Ablation studies confirm the necessity of both the semantic organizational structure and the three-tier progressive disclosure mechanism, while tool scaling experiments validate the method's scalability. As a lightweight, plug-and-play tool organization strategy, progressive disclosure offers a new perspective for agent context management in large-scale tool spaces.

\subsection{Future Work}

Situated within the broader OpenClaw-style paradigm of accomplishing tasks through dynamic tool invocation, RS-Claw demonstrates the effectiveness of progressive active tool exploration in remote sensing scenarios.  Nevertheless, further advances are still needed to make this tool-space management paradigm more adaptive, scalable, and broadly applicable.  We identify four promising directions for future research:
\begin{enumerate}
\item Exploring automatic clustering methods based on tool description semantic embeddings to replace manual categorization, and introducing adaptive split-and-merge mechanisms that allow the skill tree to automatically adjust its topology and continuously evolve as the tool library grows.

\item Constructing exploration trajectory data under the progressive disclosure interaction paradigm and optimizing the model's exploration strategy within the hierarchical structure through supervised fine-tuning or reinforcement learning.

\item Introducing an exploration budget or dynamic routing mechanism to adaptively adjust the disclosure depth based on task complexity, achieving a better trade-off between accuracy and overhead.

\item Extending the method to more vertical domains, larger-scale tool libraries, and a broader range of models for validation.
\end{enumerate}

\appendices

\section{Complete Experimental Results}\label{app:full_results}

This appendix provides complete numerical results for all experiments reported in the main text. Table~\ref{tab:full_accuracy} lists the full six-metric accuracy results for the main experiment across all three models. Table~\ref{tab:ablation_full} gives complete accuracy metrics for the ablation study. Tables~\ref{tab:scaling_same_acc} and~\ref{tab:scaling_same_ctx} report accuracy and context overhead under same-domain tool scaling. Tables~\ref{tab:scaling_cross_acc} and~\ref{tab:scaling_cross_ctx} report the corresponding results under cross-domain tool scaling.

\begin{table*}[!t]
\caption{Complete accuracy metrics for the main experiment. Six accuracy metrics for RS-Claw (active tool exploration) against two passive tool selection baselines (Flat, RAG) across three models and two evaluation modes (AP and IF), including Efficiency, Tool-Exact-Match, and Parameters not reported in the main text Table~\ref{tab:accuracy}.\label{tab:full_accuracy}}
\centering
\resizebox{\textwidth}{!}{%
\begin{tabular}{llcccccccccccc}
\toprule
\multirow{2}{*}{Model} & \multirow{2}{*}{Method} & \multicolumn{2}{c}{Accuracy} & \multicolumn{2}{c}{Efficiency} & \multicolumn{2}{c}{Tool-Any-Order} & \multicolumn{2}{c}{Tool-In-Order} & \multicolumn{2}{c}{Tool-Exact-Match} & \multicolumn{2}{c}{Parameters} \\
\cmidrule(lr){3-4}\cmidrule(lr){5-6}\cmidrule(lr){7-8}\cmidrule(lr){9-10}\cmidrule(lr){11-12}\cmidrule(lr){13-14}
& & AP & IF & AP & IF & AP & IF & AP & IF & AP & IF & AP & IF \\
\midrule
\multirow{3}{*}{GPT-5} & Flat & 65.67 & 64.38 & \textbf{2.4266} & \textbf{2.8479} & 67.40 & 69.86 & 55.89 & 58.74 & 43.58 & 44.05 & 25.84 & 25.52 \\
& RAG & 59.23 & 60.09 & 2.9558 & 4.0479 & 58.63 & 59.05 & 49.66 & 49.90 & 41.01 & 43.52 & \textbf{26.03} & \textbf{28.91} \\
& RS-Claw (Ours) & \textbf{68.67} & \textbf{70.82} & 3.3750 & 3.7426 & \textbf{73.10} & \textbf{75.57} & \textbf{59.52} & \textbf{62.63} & \textbf{50.42} & \textbf{48.37} & 24.89 & 24.64 \\
\midrule
\multirow{3}{*}{DeepSeek V3.1} & Flat & 49.36 & 51.07 & 2.6904 & 2.7142 & 77.25 & 76.66 & 61.19 & 62.19 & 46.39 & 47.75 & 30.77 & 31.40 \\
& RAG & 39.91 & 46.78 & \textbf{2.4972} & \textbf{2.1909} & 62.39 & 61.22 & 47.82 & 49.97 & 43.70 & 44.98 & 26.55 & 29.57 \\
& RS-Claw (Ours) & \textbf{57.08} & \textbf{55.79} & 2.6370 & 2.6473 & \textbf{80.77} & \textbf{81.17} & \textbf{68.62} & \textbf{69.44} & \textbf{50.48} & \textbf{52.51} & \textbf{31.37} & \textbf{32.58} \\
\midrule
\multirow{3}{*}{Qwen3-32b} & Flat & 20.60 & 25.32 & 2.8460 & 1.9208 & 39.86 & 42.36 & 22.14 & 34.31 & 9.65 & \textbf{26.15} & 8.51 & 17.85 \\
& RAG & 20.17 & 22.75 & 2.3953 & 2.3774 & 30.04 & 35.36 & 20.75 & 26.06 & \textbf{12.13} & 22.71 & \textbf{10.91} & \textbf{18.12} \\
& RS-Claw (Ours) & \textbf{33.05} & \textbf{29.18} & \textbf{1.4056} & \textbf{1.9178} & \textbf{50.22} & \textbf{60.77} & \textbf{32.06} & \textbf{44.91} & 6.11 & 20.02 & 4.97 & 11.93 \\
\bottomrule
\end{tabular}}
\end{table*}

\begin{table*}[!t]
\caption{Complete accuracy metrics for the ablation study. Six accuracy metrics for the two ablation variants (Random, 2layers) and RS-Claw under AP and IF modes, including Efficiency, Tool-Exact-Match, and Parameters not reported in the main text Table~\ref{tab:ablation_acc}.\label{tab:ablation_full}}
\centering
\resizebox{\textwidth}{!}{%
\begin{tabular}{lcccccccccccc}
\toprule
\multirow{2}{*}{Method} & \multicolumn{2}{c}{Accuracy} & \multicolumn{2}{c}{Efficiency} & \multicolumn{2}{c}{Tool-Any-Order} & \multicolumn{2}{c}{Tool-In-Order} & \multicolumn{2}{c}{Tool-Exact-Match} & \multicolumn{2}{c}{Parameters} \\
\cmidrule(lr){2-3}\cmidrule(lr){4-5}\cmidrule(lr){6-7}\cmidrule(lr){8-9}\cmidrule(lr){10-11}\cmidrule(lr){12-13}
& AP & IF & AP & IF & AP & IF & AP & IF & AP & IF & AP & IF \\
\midrule
Random & 23.18 & 25.32 & 1.9449 & 2.2380 & 44.21 & 48.13 & 31.44 & 37.93 & 5.39 & 24.93 & 4.17 & \textbf{17.44} \\
2layers & 25.75 & 25.32 & 1.8612 & 2.2380 & \textbf{58.07} & \textbf{62.50} & \textbf{43.16} & \textbf{50.37} & \textbf{17.23} & \textbf{33.35} & \textbf{10.50} & 16.65 \\
RS-Claw (Ours) & \textbf{33.05} & \textbf{29.18} & \textbf{1.4056} & \textbf{1.9178} & 50.22 & 60.77 & 32.06 & 44.91 & 6.11 & 20.02 & 4.97 & 11.93 \\
\bottomrule
\end{tabular}}
\end{table*}

\begin{table*}[!t]
\caption{Complete accuracy metrics for same-domain tool scaling. Six accuracy metrics for RS-Claw and Flat across six tool-count levels (from GT to all 104 tools) under AP and IF modes.\label{tab:scaling_same_acc}}
\centering
\resizebox{\textwidth}{!}{%
\begin{tabular}{llcccccccccccc}
\toprule
\multirow{2}{*}{Tool Count} & \multirow{2}{*}{Method} & \multicolumn{2}{c}{Accuracy} & \multicolumn{2}{c}{Efficiency} & \multicolumn{2}{c}{Tool-Any-Order} & \multicolumn{2}{c}{Tool-In-Order} & \multicolumn{2}{c}{Tool-Exact-Match} & \multicolumn{2}{c}{Parameters} \\
\cmidrule(lr){3-4}\cmidrule(lr){5-6}\cmidrule(lr){7-8}\cmidrule(lr){9-10}\cmidrule(lr){11-12}\cmidrule(lr){13-14}
& & AP & IF & AP & IF & AP & IF & AP & IF & AP & IF & AP & IF \\
\midrule
\multirow{2}{*}{GT} & Flat & \textbf{36.48} & 38.63 & 1.4988 & 1.8342 & \textbf{74.76} & \textbf{74.92} & \textbf{59.81} & \textbf{62.83} & \textbf{31.92} & \textbf{50.75} & \textbf{16.45} & \textbf{24.04} \\
& RS-Claw (Ours) & 33.48 & \textbf{39.48} & \textbf{1.2883} & \textbf{1.7317} & 60.47 & 67.25 & 40.06 & 52.42 & 5.69 & 25.39 & 3.40 & 14.22 \\
\midrule
\multirow{2}{*}{GT+20} & Flat & 28.33 & 29.61 & 2.0726 & 2.4188 & 58.20 & 62.60 & 39.65 & 50.93 & \textbf{15.18} & \textbf{33.73} & \textbf{10.76} & \textbf{20.19} \\
& RS-Claw (Ours) & \textbf{36.05} & \textbf{31.33} & \textbf{1.3579} & \textbf{1.7867} & \textbf{61.91} & \textbf{67.32} & \textbf{43.69} & \textbf{51.72} & 6.37 & 24.13 & 3.65 & 13.62 \\
\midrule
\multirow{2}{*}{GT+40} & Flat & 25.32 & 24.46 & 2.2546 & 2.6577 & \textbf{60.10} & 60.62 & \textbf{42.70} & 48.93 & \textbf{16.35} & \textbf{34.95} & \textbf{11.59} & \textbf{21.53} \\
& RS-Claw (Ours) & \textbf{27.90} & \textbf{32.62} & \textbf{1.4508} & \textbf{1.8152} & 55.62 & \textbf{64.12} & 36.97 & \textbf{49.69} & 6.48 & 21.93 & 5.58 & 12.82 \\
\midrule
\multirow{2}{*}{GT+60} & Flat & 20.17 & 22.32 & 2.4562 & 2.8187 & 50.98 & 57.37 & 35.39 & 45.81 & \textbf{14.40} & \textbf{31.47} & \textbf{10.38} & \textbf{20.55} \\
& RS-Claw (Ours) & \textbf{30.90} & \textbf{31.76} & \textbf{1.6073} & \textbf{1.9186} & \textbf{54.46} & \textbf{62.58} & \textbf{38.17} & \textbf{47.73} & 8.90 & 22.20 & 5.49 & 13.70 \\
\midrule
\multirow{2}{*}{GT+80} & Flat & 23.18 & 21.89 & 2.5908 & 2.8734 & 47.80 & 54.52 & 34.31 & 44.74 & \textbf{11.62} & \textbf{29.11} & \textbf{9.85} & \textbf{20.00} \\
& RS-Claw (Ours) & \textbf{33.05} & \textbf{30.47} & \textbf{1.3933} & \textbf{1.8616} & \textbf{50.97} & \textbf{60.54} & \textbf{34.48} & \textbf{46.29} & 6.71 & 19.76 & 3.58 & 12.29 \\
\midrule
\multirow{2}{*}{All Tools} & Flat & 20.60 & 25.32 & 2.8460 & 1.9208 & 39.86 & 42.36 & 22.14 & 34.31 & \textbf{9.65} & \textbf{26.15} & \textbf{8.51} & \textbf{17.85} \\
& RS-Claw (Ours) & \textbf{33.05} & \textbf{29.18} & \textbf{1.4056} & \textbf{1.9178} & \textbf{50.22} & \textbf{60.77} & \textbf{32.06} & \textbf{44.91} & 6.11 & 20.02 & 4.97 & 11.93 \\
\bottomrule
\end{tabular}}
\end{table*}

\begin{table*}[!t]
\caption{Context overhead metrics for same-domain tool scaling. Average input tokens per question and per turn for RS-Claw and Flat across six tool-count levels under AP and IF modes.\label{tab:scaling_same_ctx}}
\centering
\begin{tabular}{llcccc}
\toprule
\multirow{2}{*}{Tool Count} & \multirow{2}{*}{Method} & \multicolumn{2}{c}{Tokens for each question} & \multicolumn{2}{c}{Tokens for each turn} \\
\cmidrule(lr){3-4}\cmidrule(lr){5-6}
& & AP & IF & AP & IF \\
\midrule
\multirow{2}{*}{GT} & Flat & \textbf{41,520} & \textbf{54,797} & 4,876 & 5,316 \\
& RS-Claw (Ours) & 45,321 & 75,231 & \textbf{4,080} & \textbf{5,202} \\
\midrule
\multirow{2}{*}{GT+20} & Flat & 108,492 & 143,138 & 9,949 & 11,137 \\
& RS-Claw (Ours) & \textbf{44,319} & \textbf{82,773} & \textbf{3,844} & \textbf{5,733} \\
\midrule
\multirow{2}{*}{GT+40} & Flat & 191,689 & 231,905 & 15,997 & 16,395 \\
& RS-Claw (Ours) & \textbf{59,941} & \textbf{81,195} & \textbf{4,940} & \textbf{5,512} \\
\midrule
\multirow{2}{*}{GT+60} & Flat & 255,052 & 316,105 & 20,408 & 21,565 \\
& RS-Claw (Ours) & \textbf{68,153} & \textbf{96,171} & \textbf{5,023} & \textbf{6,050} \\
\midrule
\multirow{2}{*}{GT+80} & Flat & 351,563 & 407,993 & 26,353 & 27,469 \\
& RS-Claw (Ours) & \textbf{59,719} & \textbf{97,992} & \textbf{4,890} & \textbf{6,297} \\
\midrule
\multirow{2}{*}{All Tools} & Flat & 502,119 & 517,445 & 30,612 & 31,756 \\
& RS-Claw (Ours) & \textbf{70,759} & \textbf{97,325} & \textbf{5,951} & \textbf{6,309} \\
\bottomrule
\end{tabular}
\end{table*}

\begin{table*}[!t]
\caption{Complete accuracy metrics for cross-domain tool scaling. Six accuracy metrics for RS-Claw and Flat across three cross-domain expansion stages (104$\to$179$\to$234 tools) under AP and IF modes.\label{tab:scaling_cross_acc}}
\centering
\resizebox{\textwidth}{!}{%
\begin{tabular}{llcccccccccccc}
\toprule
\multirow{2}{*}{Tool Count} & \multirow{2}{*}{Method} & \multicolumn{2}{c}{Accuracy} & \multicolumn{2}{c}{Efficiency} & \multicolumn{2}{c}{Tool-Any-Order} & \multicolumn{2}{c}{Tool-In-Order} & \multicolumn{2}{c}{Tool-Exact-Match} & \multicolumn{2}{c}{Parameters} \\
\cmidrule(lr){3-4}\cmidrule(lr){5-6}\cmidrule(lr){7-8}\cmidrule(lr){9-10}\cmidrule(lr){11-12}\cmidrule(lr){13-14}
& & AP & IF & AP & IF & AP & IF & AP & IF & AP & IF & AP & IF \\
\midrule
\multirow{2}{*}{RS-Claw (104)} & Flat & 20.60 & 25.32 & 2.8460 & 1.9208 & 39.86 & 42.36 & 22.14 & 34.31 & \textbf{9.65} & \textbf{26.15} & \textbf{8.51} & \textbf{17.85} \\
& RS-Claw (Ours) & \textbf{33.05} & \textbf{29.18} & \textbf{1.4056} & \textbf{1.9178} & \textbf{50.22} & \textbf{60.77} & \textbf{32.06} & \textbf{44.91} & 6.11 & 20.02 & 4.97 & 11.93 \\
\midrule
\multirow{2}{*}{stage1 (179)} & Flat & 20.17 & 21.46 & 2.2769 & 2.4426 & 42.55 & 52.60 & 25.01 & 42.42 & \textbf{6.61} & \textbf{28.28} & \textbf{6.18} & \textbf{18.55} \\
& RS-Claw (Ours) & \textbf{28.33} & \textbf{29.18} & \textbf{1.4794} & \textbf{2.0007} & \textbf{53.17} & \textbf{59.31} & \textbf{36.29} & \textbf{45.61} & 1.80 & 18.70 & 1.00 & 12.83 \\
\midrule
\multirow{2}{*}{stage2 (234)} & Flat & 21.46 & 24.89 & 2.5781 & 2.7434 & 43.52 & 53.08 & 27.52 & 42.56 & \textbf{11.53} & \textbf{27.63} & \textbf{9.05} & \textbf{18.53} \\
& RS-Claw (Ours) & \textbf{28.33} & \textbf{30.47} & \textbf{1.4136} & \textbf{1.7855} & \textbf{50.14} & \textbf{59.94} & \textbf{32.46} & \textbf{44.30} & 1.01 & 18.54 & 0.59 & 9.47 \\
\bottomrule
\end{tabular}}
\end{table*}

\begin{table*}[!t]
\caption{Context overhead metrics for cross-domain tool scaling. Average input tokens per question and per turn for RS-Claw and Flat across three cross-domain expansion stages under AP and IF modes.\label{tab:scaling_cross_ctx}}
\centering
\begin{tabular}{llcccc}
\toprule
\multirow{2}{*}{Tool Count} & \multirow{2}{*}{Method} & \multicolumn{2}{c}{Tokens for each question} & \multicolumn{2}{c}{Tokens for each turn} \\
\cmidrule(lr){3-4}\cmidrule(lr){5-6}
& & AP & IF & AP & IF \\
\midrule
\multirow{2}{*}{RS-Claw (104)} & Flat & 502,119 & 517,445 & 30,612 & 31,756 \\
& RS-Claw (Ours) & \textbf{70,759} & \textbf{97,325} & \textbf{5,951} & \textbf{6,309} \\
\midrule
\multirow{2}{*}{stage1 (179)} & Flat & 467,852 & 528,073 & 39,861 & 40,996 \\
& RS-Claw (Ours) & \textbf{62,768} & \textbf{110,011} & \textbf{5,093} & \textbf{6,814} \\
\midrule
\multirow{2}{*}{stage2 (234)} & Flat & 642,438 & 697,470 & 49,046 & 50,193 \\
& RS-Claw (Ours) & \textbf{72,156} & \textbf{113,016} & \textbf{5,947} & \textbf{7,275} \\
\bottomrule
\end{tabular}
\end{table*}

\section{System Prompt and Skill Tree Design}

\subsection{System Prompt Template}

The following is the complete system prompt used by RS-Claw. The \texttt{\{kit\_table\}} placeholder is dynamically populated at runtime with a JSON block listing the five skill kits and their applicable task descriptions (see Section~A.2.2).

\begin{rsclawbox}[RS-Claw system prompt]
You are a geoscientist, and you need to use tools to answer multiple-choice questions about Earth observation data analysis. Note that if a tool returns an error, you can only try again once. Ultimately, you only need to explicitly tell me the correct choice.

ATTENTION:
1. When a tool returns "Result saved at /path/to/file", you must use the full returned path "/path/to/file" in all subsequent tool calls.
2. For each question, you must provide the choice you think is most appropriate. Your final answer format must be: <Answer>Your choice<Answer>
3. You have access to EO tools via a 4-meta-tool progressive disclosure interface: skill, doc, call, filelist. You MUST follow the strict order: skill -> doc -> call. Skipping any step is forbidden. If a tool call returns an error, re-read the doc output carefully and fix the parameters. Only retry once -- if it fails again, try a different tool or approach. For doc and call, use tool_id in 'kit.tool_name' format, e.g. 'statistics.calc_batch_image_mean'. For call, tool_args must be a JSON string, e.g. tool_args='{"file_list": ["a.tif", "b.tif"]}' Available kits and their applicable tasks:
{kit_table}
\end{rsclawbox}

The \texttt{\{kit\_table\}} block is rendered as:

\begin{rsclawbox}[{kit\_table} rendered value]
[
  {
    "kit": "index",
    "applicable_tasks": "Remote sensing index calculation for vegetation, water body, building, snow/ice, fire, etc.",
    "typical_usage": "Calculate NDVI, NDWI, TVDI and other indices, results saved as .tif files"
  },
  {
    "kit": "inversion",
    "applicable_tasks": "Only for inverting physical quantities from raw bands (LST, water vapor, sea ice concentration, water turbidity, etc.)",
    "typical_usage": "Input raw bands, output physical quantity .tif files; not for statistical analysis"
  },
  {
    "kit": "perception",
    "applicable_tasks": "Visual perception tasks including image segmentation, object detection, change detection, scene classification",
    "typical_usage": "Use when identifying targets or change areas in images"
  },
  {
    "kit": "analysis",
    "applicable_tasks": "Time series analysis (trend, change points, seasonal decomposition, hotspot spatial analysis)",
    "typical_usage": "Input numeric sequences for trend/change point/seasonality analysis; input rasters for spatial hotspot analysis"
  },
  {
    "kit": "statistics",
    "applicable_tasks": "First choice for area/pixel percentage calculation; image statistics, math operations, temperature conversion, fire analysis",
    "typical_usage": "Use calculate_threshold_ratio for above-threshold pixel percentage; use calc_batch_image_mean for batch mean values"
  }
]
\end{rsclawbox}

This skill summary constitutes the skill layer $\{\sigma(s_m)\}_{m=1}^{M}$ of the hierarchical skill tree, providing the agent with coarse-grained navigation priors before any tool-level information is loaded.

\subsection{Skill Tree Structure}\label{app:skill_tree}

The five skill nodes and their constituent tools are listed below. Each table corresponds to the tool catalog layer $\{\psi(t)\}_{t \in \mathcal{T}_m}$ returned when the agent calls \texttt{skill(kit)}. Detailed execution documents $d_t$ are only revealed upon a subsequent \texttt{doc(kit.tool\_name)} call.

\begin{kitbox}[Skill Tree: Five Kits and Constituent Tools (104 tools)]
Index Kit (12 tools)
  calculate_batch_ndvi
  calculate_batch_ndwi
  calculate_batch_ndbi
  calculate_batch_evi
  calculate_batch_nbr
  calculate_batch_fvc
  calculate_batch_wri
  calculate_batch_ndti
  calculate_batch_frp
  calculate_batch_ndsi
  calc_extreme_snow_loss_percentage_from_binary_map
  compute_tvdi

Inversion Kit (17 tools)
  lst_single_channel
  lst_multi_channel
  split_window
  ttm_lst
  modis_day_night_lst
  temperature_emissivity_separation
  ATI
  calculate_mean_lst_by_ndvi
  calculate_max_lst_by_ndvi
  band_ratio
  dual_polarization_differential
  dual_frequency_diff
  multi_freq_bt
  chang_single_param_inversion
  nasa_team_sea_ice_concentration
  dual_polarization_ratio
  calculate_water_turbidity_ntu

Perception Kit (15 tools)
  threshold_segmentation
  count_above_threshold
  count_skeleton_contours
  bbox_expansion
  bboxes2centroids
  centroid_distance_extremes
  calculate_bbox_area
  MSCN
  RemoteCLIP
  Strip_R_CNN
  SM3Det
  RemoteSAM
  InstructSAM
  SAM2
  ChangeOS

Analysis Kit (10 tools)
  compute_linear_trend
  mann_kendall_test
  sens_slope
  stl_decompose
  detect_change_points
  autocorrelation_function
  detect_seasonality_acf
  count_spikes_from_values
  getis_ord_gi_star
  analyze_hotspot_direction

Statistics Kit (50 tools)
  calc_batch_image_mean
  calc_batch_image_std
  calc_batch_image_median
  calc_batch_image_min
  calc_batch_image_max
  calc_batch_image_skewness
  calc_batch_image_kurtosis
  calc_batch_image_sum
  calc_batch_image_hotspot_percentage
  calc_batch_image_hotspot_tif
  calc_batch_image_mean_mean
  calc_batch_image_mean_max
  calc_batch_image_mean_max_min
  calc_batch_image_mean_threshold
  difference
  division
  multiply
  subtract
  percentage_change
  ceil_number
  mean
  max_value_and_index
  min_value_and_index
  get_list_object_via_indexes
  kelvin_to_celsius
  celsius_to_kelvin
  calc_batch_fire_pixels
  create_fire_increase_map
  identify_fire_prone_areas
  get_percentile_value_from_image
  calculate_area
  calculate_threshold_ratio
  calculate_multi_band_threshold_ratio
  count_pixels_satisfying_conditions
  count_images_exceeding_threshold_ratio
  average_ratio_exceeding_threshold
  count_images_exceeding_mean_multiplier
  calculate_band_mean_by_condition
  calc_threshold_value_mean
  coefficient_of_variation
  skewness
  kurtosis
  image_division_mean
  calculate_intersection_percentage
  calculate_tif_average
  calculate_tif_difference
  grayscale_to_colormap
  get_filelist
  radiometric_correction_sr
  apply_cloud_mask
\end{kitbox}

\section{Implementation Details}

\subsection{Baseline Implementation}

Both the Flat and RAG baselines share the same system prompt, which contains only task instructions without any skill-level navigation information:

\begin{promptbox}[Flat / RAG baseline system prompt]
You are a geoscientist, and you need to use tools to answer multiple-choice questions about Earth observation data analysis. Note that if a tool returns an error, you can only try again once. Ultimately, you only need to explicitly tell me the correct choice.

ATTENTION:
1. When a tool returns "Result saved at /path/to/file", you must use the full returned path "/path/to/file" in all subsequent tool calls.
2. For each question, you must provide the choice you think is most appropriate. Your final answer format must be: <Answer>Your choice<Answer>
\end{promptbox}

\textbf{Flat baseline.} The Flat baseline follows the official Earth-Agent implementation. All 104 domain-specific tools are exposed to the agent via the Model Context Protocol (MCP). At agent initialization, the MCP client connects to the tool servers and loads the complete tool list; each tool's name and full description are automatically injected into the LLM's context as part of the function-calling schema. The system prompt contains only task instructions with no skill-level navigation information. The agent directly selects and invokes tools from the full flat list in each reasoning step.

\textbf{RAG baseline.} The RAG baseline uses a retrieval-augmented tool selection strategy. All 104 tools are loaded from the MCP servers at startup. A FAISS vector index is built offline over tool descriptions using a local Ollama embedding model (\texttt{nomic-embed-text}). For each question, the query text is embedded and used to retrieve the top-$k$ most semantically similar tools ($k = 19$). The tool \texttt{get\_filelist} is always force-included regardless of retrieval score, yielding a fixed context of 20 tools per question. A fresh ReAct agent is instantiated with only the retrieved tool subset for each question.

\subsection{Evaluation Metric Computation}

\textbf{End-to-end metrics.} Accuracy is computed as the fraction of questions for which the agent's final answer matches the ground-truth choice. The agent's answer is extracted from the response using the \texttt{<Answer>X<Answer>} or \texttt{<Answer>X</Answer>} tag pattern. Efficiency is defined as the ratio of the number of tool calls made by the agent to the number of tool calls in the ground-truth trajectory: $\text{Efficiency} = N_{\text{model}} / N_{\text{GT}}$. A value greater than 1 indicates the agent used more tool calls than the reference solution.

\textbf{Step-level metrics.} Four step-level metrics are computed by comparing the agent's tool call sequence against the ground-truth trajectory:
\begin{itemize}
\item Tool-Any-Order: the fraction of ground-truth tools that appear anywhere in the agent's tool call sequence, regardless of order. Formally, $|\mathcal{T}_{\text{pred}} \cap \mathcal{T}_{\text{GT}}| / |\mathcal{T}_{\text{GT}}|$, where sets are used (duplicates collapsed).
\item Tool-In-Order: the fraction of ground-truth tools matched in order, allowing intervening tool calls. A greedy sequential scan is used: for each expected tool in order, the earliest remaining occurrence in the predicted sequence is consumed.
\item Tool-Exact-Match: the fraction of positions where the predicted and ground-truth tool sequences agree exactly, up to the length of the shorter sequence. Matching stops at the first mismatch.
\item Parameters: the fraction of ground-truth steps for which both the tool name and all input parameters exactly match the prediction. Matching proceeds step by step and stops at the first mismatch.
\end{itemize}
All four metrics are soft-scored (partial credit per question) and averaged across questions.

\textbf{Context overhead metrics.} For each agent turn, the number of input (prompt) tokens is recorded from the LLM response metadata. Two aggregate statistics are reported:
\begin{enumerate}
\item tokens per question---the sum of prompt tokens across all turns for a given question, averaged over all questions;
\item tokens per turn---the per-turn prompt token count, averaged over all turns across all questions.
\end{enumerate}

\section{Case Study}

\subsection{Comparative Trajectory Cases Against Flat}\label{app:case_flat}

This subsection presents two representative cases to illustrate the effectiveness of the progressive skill tree disclosure strategy at the trajectory level. The two cases correspond respectively to the two core problems identified in the introduction: tool hallucination (Case F1) and restricted reasoning space (Case F2). In each case, RS-Claw arrives at the correct answer through actual computation, while the Flat baseline fails or produces a wrong answer on the same task.

\subsubsection{Case F1 --- ATI Drought Stress Analysis}

\textbf{Task:} Using satellite thermal bands and surface albedo to compute ATI, calculate the proportion of the Sahel region with ATI $< 1.0$ indicating drought stress in May 2023. Correct answer: D (70.92\%).

\begin{rsclawbox}[RS-Claw trajectory (DeepSeek-V3.1{,} AP)]
[Explore] skill("inversion")
              -> sees: ATI -- Apparent Thermal Inertia inversion
[Explore] doc("inversion.ATI")
[Explore] skill("statistics")
              -> sees: calculate_threshold_ratio -- pixel % above/below threshold
[Explore] doc("statistics.calculate_threshold_ratio")
[Execute] filelist(question42/)
[Execute] call inversion.ATI x10 (one per day, May 1-10)
              -> Result saved at .../ati_result_*.tif x10
[Execute] call statistics.calculate_threshold_ratio(
              image_paths=[...x10], threshold=1.0)
              -> 70.92%
Answer: D (correct)
\end{rsclawbox}

\begin{promptbox}[Flat trajectory (DeepSeek-V3.1{,} AP)]
[Execute] get_filelist
[Execute] calc_batch_image_mean_mean  -> FAIL (wrong tool, pre-ATI)
[Execute] ATI x1 (one day only)
[Execute] calc_batch_image_hotspot_percentage -> 0.55% (wrong metric)
[Execute] calculate_threshold_ratio -> FAIL (parameter error)
[Execute] calculate_tif_average -> FAIL (broadcast error)
[Execute] threshold_segmentation -> FAIL (nodata error)
[Execute] calculate_area / count_above_threshold / division / ...  x6
              -> 25339 / 21286 / 0.84 / ...  (meaningless intermediates)
Answer: FAIL (incorrect)
\end{promptbox}

RS-Claw navigates two skill layers (\texttt{skill("inversion")} $\to$ \texttt{skill("statistics")}) to precisely narrow the visible tools to ATI and \texttt{calculate\_threshold\_ratio}, forming a clean two-step chain and arriving at the correct answer. The Flat baseline, operating under the full context of 104 tools, is misled after the ATI call by 13 semantically adjacent but functionally irrelevant tools, resulting in repeated failures and ultimately FAIL. This directly illustrates tool hallucination: injecting the full toolset causes the model to confuse semantically similar tools, while the skill tree's hierarchical filtering constrains visible tool information to the locally relevant subset.

\subsubsection{Case F2 --- Dead Sea NDTI Turbidity Change Analysis}

\textbf{Task:} Using reflectance data of the Dead Sea for August 2020 and 2022, compute NDTI and analyze the turbidity change trend. Correct answer: B (change $+0.66063$, increasing).

\begin{rsclawbox}[RS-Claw trajectory (GPT-5{,} AP)]
[Execute] filelist(question141/)
[Explore] skill("index")
              -> sees: calculate_batch_ndti -- NDTI from Red/Green bands
[Explore] doc("index.calculate_batch_ndti")
[Explore] skill("statistics")
              -> sees: calc_batch_image_mean_mean -- mean of means across batch
[Explore] doc("statistics.calc_batch_image_mean_mean")
[Execute] call index.calculate_batch_ndti(2020 files)
              -> ndti_2020_*.tif x15
[Execute] call statistics.calc_batch_image_mean_mean(
              [ndti_2020_*.tif]) -> 121191.78
[Execute] call index.calculate_batch_ndti(2022 files)
              -> ndti_2022_*.tif x15
[Execute] call statistics.calc_batch_image_mean_mean(
              [ndti_2022_*.tif]) -> 55128.48
              diff = 55128.48 - 121191.78 = -66063.30
              -> turbidity increased
Answer: B (correct)
\end{rsclawbox}

\begin{promptbox}[Flat trajectory (GPT-5{,} AP)]
[Execute] get_filelist
[Execute] calculate_batch_ndti(all files) -> ndti_*.tif x30
[Execute] calc_batch_image_sum([ndti_*.tif])
              -> [-360000192, -200001184, ...] (meaningless raw sums)
[Execute] count_above_threshold x4 -> [1272, 1516, ...] (pixel counts)
[Execute] mean([1272, 1516, ...]) -> 1226 / 1114 (wrong aggregation)
[Execute] calculate_tif_average / calculate_tif_difference
[Execute] count_pixels_satisfying_conditions -> 1092
[Execute] calculate_band_mean_by_condition -> 291904.22
[Execute] subtract / calc_batch_image_sum x2  -> -321400576 / -146200736
[Execute] mean(...) -> 1461.3
Answer: A (incorrect)
\end{promptbox}

Both agents correctly identify \texttt{calculate\_batch\_ndti}, but their subsequent paths diverge sharply. RS-Claw completes the task in 8 steps with 4 distinct tools, maintaining clear intermediate states throughout (means 121191.78 / 55128.48, difference directly matching the answer). The Flat baseline, under the context load of the full toolset, becomes severely overloaded: after the NDTI computation it attempts 11 additional tools, generating a flood of meaningless intermediates (pixel counts 1226/1114, sum $-$360,000,192, etc.) that overwhelm the reasoning, ultimately producing the wrong answer A. The failure is not a tool selection error but a constrained reasoning space problem: full-toolset registration prevents the agent from maintaining coherent intermediate states amid the noise---a micro-level corroboration of the quantitative finding that RS-Claw incurs substantially lower per-turn token overhead than Flat.

\subsection{Ablation Evidence Cases: RS-Claw vs. 2layers}\label{app:case_2layers}

In the ablation study, 2layers achieves a higher overall Tool-Any-Order (TAO, 58.07) than RS-Claw (50.22), confirming that pre-exposing all tool names does improve tool discovery coverage. Yet 2layers' accuracy (25.75\%) falls below RS-Claw's (33.05\%). In both cases below, Tool-Any-Order (TAO) and Tool-In-Order (TIO) scores are identical across the two methods, ruling out any difference in tool discovery ability and attributing failure directly to constrained reasoning space after omitting the skill summary layer: under Qwen3-32b, greater tool visibility does not translate into higher accuracy---reasoning space is the decisive bottleneck.

\subsubsection{Case A1 --- Split-Window LST High-Temperature Ratio}

\textbf{Task:} Using thermal Band 31 and 32 data from irrigated farmland in northern Hebei on August 5, 2021, apply the split-window algorithm to compute LST, then calculate the percentage of high-temperature pixels ($>305$ K). Correct answer: D (63.17\%).

\begin{rsclawbox}[RS-Claw trajectory (Qwen3-32b{,} IF{,} TAO{=}1.00{,} TIO{=}1.00)]
[Explore] skill("inversion")
              -> sees: split_window -- LST from Band 31/32
[Explore] doc("inversion.split_window")
[Execute] call inversion.split_window(b31.tif) -> FAIL (wrong filename)
[Execute] filelist(question33/)
[Execute] call inversion.split_window(2021_08_05_0310_BT_31.tif, BT_32.tif)
              -> Result saved at .../lst_result.tif
[Explore] doc("statistics.calculate_threshold_ratio")
[Execute] call statistics.calculate_threshold_ratio(lst_result.tif, threshold=305)
              -> 63.17%
Answer: D (correct, 20,415 tokens)
\end{rsclawbox}

\begin{promptbox}[2layers trajectory (Qwen3-32b{,} IF{,} TAO{=}1.00{,} TIO{=}1.00)]
[Explore] doc(split_window)
[Execute] call split_window(b31.tif) -> FAIL (wrong filename)
[Execute] filelist(question33/)
[Execute] call split_window(2021_08_05_0310_BT_31.tif, BT_32.tif)
              -> Result saved at .../lst_output.tif
[Explore] doc(threshold_segmentation) <- distracted by adjacent tool in system prompt
[Execute] call threshold_segmentation(lst_output.tif, threshold=305)
              -> mask saved
[Explore] doc(calculate_multi_band_threshold_ratio)
[Execute] call calculate_multi_band_threshold_ratio(lst_output.tif, [[0,305,"above"]])
              -> FAIL (output_path required)
[Explore] doc(calculate_multi_band_threshold_ratio)
[Execute] call calculate_multi_band_threshold_ratio(lst_output.tif, [[0,305,"above"]])
              -> FAIL (same error)
[Explore] doc(threshold_segmentation) <- back to threshold_segmentation again
[Execute] call threshold_segmentation(BT_31.tif, threshold=305) x4
              -> mask saved x4 (operating on raw BT file instead of LST result)
[Explore] doc(calculate_threshold_ratio)
[Execute] call calculate_threshold_ratio(BT_31.tif, threshold=305, mode="above")
              -> 47.04% (wrong file, answer too low)
Answer: C (incorrect, 94,715 tokens)
\end{promptbox}

Both TAO and TIO are 1.00 for both methods, confirming that 2layers correctly identified all ground-truth tools---tool visibility is not the issue. However, after successfully running \texttt{split\_window} to produce \texttt{lst\_output.tif}, 2layers was drawn to semantically adjacent tools pre-exposed in the system prompt (\texttt{threshold\_segmentation}, \texttt{calculate\_multi\_band\_threshold\_ratio}), repeatedly consulting their docs and calling them, generating a large volume of spurious intermediate steps. Under the context load of 94,715 tokens, the model ultimately ran \texttt{calculate\_threshold\_ratio} on the raw BT file rather than the LST result, yielding the incorrect 47.04\%. RS-Claw confined visible tools to the \texttt{inversion} and \texttt{statistics} scopes via the skill layer, keeping context clean throughout, and completed correct reasoning in just 16 turns and 20,415 tokens. This case directly demonstrates that constrained reasoning space alone is sufficient to cause the model to confuse intermediate files at the aggregation stage and produce a wrong answer.

\subsubsection{Case A2 --- MODIS Atmospheric Water Vapor Monthly Mean}

\textbf{Task:} Using MODIS bands b02, b05, b17, b18, b19 surface reflectance data over the Turpan region in July 2020, estimate daily atmospheric water vapor via the band ratio method and compute the monthly average. Correct answer: C (11.3910).

\begin{rsclawbox}[RS-Claw trajectory (Qwen3-32b{,} AP{,} TAO{=}0.67{,} TIO{=}0.67)]
[Explore] skill("inversion")
              -> sees: band_ratio -- atmospheric water vapor from MODIS bands
[Explore] doc("inversion.band_ratio")
[Execute] call inversion.band_ratio(b02.tif, ...) -> FAIL (wrong path)
[Execute] filelist(".")
[Execute] call inversion.band_ratio(question61/b02.tif, ...) -> FAIL (wrong path)
[Execute] filelist(benchmark/data/question61/)
[Execute] call inversion.band_ratio(2020_07_01_0455_Reflectance_2.tif, ...)
              -> pwv_2020_07_01.tif (9.615 g/cm2)
[Execute] call inversion.band_ratio(2020_07_15_0505_Reflectance_2.tif, ...)
              -> pwv_2020_07_15.tif (12.234 g/cm2)
[Execute] call inversion.band_ratio(2020_07_29_0515_Reflectance_2.tif, ...)
              -> pwv_2020_07_29.tif (12.324 g/cm2)
[Explore] skill("statistics")
[Explore] doc("statistics.calc_batch_image_mean")
[Execute] call statistics.calc_batch_image_mean([pwv_07_01, pwv_07_15, pwv_07_29])
              -> [9.615, 12.234, 12.324]
[Execute] call statistics.mean([9.615, 12.234, 12.324])
              -> 11.391
Answer: C (correct, 66,366 tokens)
\end{rsclawbox}

\begin{promptbox}[2layers trajectory (Qwen3-32b{,} AP{,} TAO{=}0.67{,} TIO{=}0.67)]
[Explore] doc(band_ratio)
[Execute] call band_ratio(b02.tif, ...) -> FAIL (wrong path)
[Execute] filelist(benchmark/data/question61/)
[Execute] call band_ratio(2020_07_01_0455_Reflectance_2.tif, ...)
              -> pwv_2020_07_01_0455.tif (9.615 g/cm2)
              <- only July 1st computed, remaining dates not iterated
[Execute] call calc_batch_image_mean([pwv_2020_07_01_0455.tif])
              -> 9.615 (single-day value used as monthly mean)
Answer: D (incorrect, 20,992 tokens)
\end{promptbox}

Both TAO and TIO are 0.67 for both methods, confirming that 2layers correctly identified the two core tools (\texttt{band\_ratio} and \texttt{calc\_batch\_image\_mean})---tool discovery ability is on par with RS-Claw. However, in AP mode, the 104 tool names pre-exposed in 2layers' system prompt consumed substantial context, leaving the model insufficient reasoning space (20,992 tokens) to complete the critical planning step of iterating over all dates in July. It ran \texttt{band\_ratio} for July 1st only, then immediately passed the single-day result 9.615 as the monthly mean, selecting D. RS-Claw loaded tool information on demand via the skill layer, preserving ample reasoning space, and autonomously planned the full ``per-date computation $\to$ average'' pipeline in AP mode, correctly calling \texttt{band\_ratio} for all three dates before averaging to 11.391. This case demonstrates that in AP autonomous-planning mode, insufficient reasoning space directly truncates multi-step planning depth---the tools were found, but the task was prematurely closed.

\bibliographystyle{IEEEtran}
\bibliography{refs}

\end{document}